\documentclass{article} 
\usepackage{nips15submit_e,times}
\usepackage[numbers]{natbib}
\usepackage{comment}
\usepackage{url}
\usepackage{epsfig}
\usepackage{boxedminipage}
\usepackage{wrapfig}
\usepackage{amsmath,amssymb}
\usepackage{algorithm}
\usepackage{times}
\usepackage{amsmath}
\usepackage{amsthm}
\usepackage{amssymb}
\usepackage{graphicx} 
\usepackage{enumitem}
\usepackage{graphicx}
\usepackage{booktabs}
\usepackage{xcolor}
\usepackage{float}
\usepackage{wrapfig}
\usepackage{algpseudocode}
\usepackage{subcaption}
\usepackage[font={small}]{caption}
\usepackage{textcomp}

\newcommand\AAAA{Feedforward}
\newcommand\BBBB{Recurrent}
\newcommand\A{\AAAA{} encoding}
\newcommand\B{\BBBB{} encoding}
\newcommand\AAA{feedforward encoding}
\newcommand\BBB{recurrent encoding}

\usepackage[compact]{titlesec}
\titlespacing*{\paragraph}{0ex}{0ex minus .1 ex}{1pc}
\titlespacing{\subsection}{0pt}{*0.5}{*0}
\titlespacing{\section}{0pt}{*1.2}{*0}

\usepackage[pagebackref=false,breaklinks=true,colorlinks,bookmarks=false,citecolor=blue]{hyperref}

\usepackage{color}
\definecolor{gray}{rgb}{0.5,0.5,0.5}

\iffalse 

        \newcommand{\TODO}[1]{}
        \newcommand{\outline}[1]{}
        \newcommand{\textgray}[1]{}
        \newcommand{\commenttext}[1]{}
        \newcommand{\commentfoot}[1]{}
        \newcommand{\commentselfoot}[2]{}
        \newcommand{\commentselrep}[2]{}
        \newcommand{\topic}[1]{}
\else 
        \newcommand{\TODO}[1]{{\textcolor{red}{[[TODO: {#1}]]}}}
        \newcommand{\outline}[1]{{\textcolor{blue}{[[{#1}]]}}}
        \newcommand{\textgray}[1]{\textcolor{gray}{[[{#1}]]}}
        \newcommand{\commenttext}[1]{\textcolor{red}{[[{#1}]]}}
        \newcommand{\commentfoot}[1]{\footnote{\textcolor{red}{\textit{#1}}}}
        \newcommand{\commentselfoot}[2]{{\textcolor{blue}{#1}}\commenttext{#2}}
        \newcommand{\commentselrep}[2] {{\textcolor{blue}{#1}} {\textcolor{green}{[[\textit{#2}]]}}}
        \newcommand{\topic}[1]{\textcolor{gray}{\textbf{(#1.)}}}
\fi

\newcommand{\cutabstractup}{\vspace*{-0.25in}}
\newcommand{\cutabstractdown}{\vspace*{-0.1in}}

\newcommand{\cutsectionup}{\vspace*{-0.08in}}
\newcommand{\cutsectiondown}{\vspace*{-0.04in}}

\newcommand{\cutsubsectionup}{\vspace*{-0.04in}} 
\newcommand{\cutsubsectiondown}{\vspace*{-0.04in}} 

\newcommand{\cutparagraphup}{\vspace*{-0.02in}}

\title{Action-Conditional Video Prediction \\ using Deep Networks in Atari Games}

\author{
Junhyuk Oh \texttt{  }\texttt{  }
Xiaoxiao Guo\texttt{  }\texttt{  }
Honglak Lee\texttt{  }\texttt{  }
Richard Lewis \texttt{  }\texttt{  }
Satinder Singh \\
University of Michigan, Ann Arbor, MI 48109, USA\\
\texttt{\{junhyuk,guoxiao,honglak,rickl,baveja\}}@umich.edu
}

\nipsfinalcopy 

\begin{document}
\maketitle

\cutabstractup
\begin{abstract}
\cutabstractdown
Motivated by vision-based reinforcement learning (RL) problems, in particular Atari games from the recent benchmark Aracade Learning Environment (ALE), we consider spatio-temporal prediction problems where future image-frames depend on control variables or actions as well as previous frames. 
While not composed of natural scenes, frames in Atari games are high-dimensional in size, can involve tens of objects with one or more objects being controlled by the actions directly and many other objects being influenced indirectly, can involve entry and departure of objects, and can involve deep partial observability. 
We propose and evaluate two deep neural network architectures that consist of encoding, action-conditional transformation, and decoding layers based on convolutional neural networks and recurrent neural networks.
Experimental results show that the proposed architectures are able to generate visually-realistic frames that are also useful for control over approximately 100-step action-conditional futures in some games. 
To the best of our knowledge, this paper is the first to make and evaluate long-term predictions on high-dimensional video conditioned by control inputs. 
\end{abstract}
\vspace*{-0.05in}

\cutsectionup
\section{Introduction}
\cutsectiondown
Over the years, deep learning approaches (see~\cite{bengio2009learning,schmidhuber2015deep} for survey) have shown great success in many visual perception problems (e.g.,~\cite{krizhevsky2012imagenet,ciresan2012multi,szegedy2014going,girshick2014rich}). However, modeling videos (building a generative model) is still a very challenging problem because it often involves high-dimensional natural-scene data with complex temporal dynamics.
Thus, recent studies have mostly focused on modeling simple video data, such as bouncing balls or small patches, where the next frame is highly-predictable given the previous frames~\cite{sutskever2009recurrent,mittelman2014structured,michalski2014modeling}. 
In many applications, however, future frames depend not only on previous frames but also on  control or action variables. 
For example, the first-person-view in a vehicle is affected by wheel-steering and acceleration. 
The camera observation of a robot is similarly dependent on its movement and changes of its camera angle. More generally, in vision-based reinforcement learning (RL) problems, learning to predict future images conditioned on actions amounts to learning a model of the dynamics of the agent-environment interaction,
an essential component of model-based
approaches to RL. In this paper, we focus on Atari games from the Arcade Learning Environment (ALE)~\cite{bellemare13arcade} as a source of challenging action-conditional video modeling problems. While not composed of natural scenes, frames in Atari games are high-dimensional, can involve tens of objects with one or more objects being controlled by the actions directly and many other objects being influenced indirectly, can involve entry and departure of objects, and can involve deep partial observability. To the best of our knowledge, this paper is the first to make and evaluate long-term predictions on high-dimensional images conditioned by control inputs. 

This paper proposes, evaluates, and contrasts two spatio-temporal prediction architectures based on deep networks that incorporate action variables (See Figure~\ref{fig:arch}). 
Our experimental results show that our architectures are able to generate realistic frames over 100-step action-conditional future frames without diverging in some Atari games. We show that the representations learned by our architectures 1) approximately capture natural similarity among actions, and 2) discover which objects are directly controlled by the agent's actions and which are only indirectly influenced or not controlled.
We evaluated the usefulness of our architectures for control in two ways: 1) by replacing emulator frames with predicted frames in a previously-learned model-free controller (DQN; DeepMind's state of the art Deep-Q-Network for Atari Games~\cite{mnih2013playing}), and 2) by using the predicted frames to drive a more informed than random exploration strategy to improve a model-free controller (also DQN).

\cutsectionup 
\section{Related Work}

\cutparagraphup
\paragraph{Video Prediction using Deep Networks.} 
The problem of video prediction has led to a variety of architectures in deep learning. 
A recurrent temporal restricted Boltzmann machine (RTRBM)~\cite{sutskever2009recurrent} was proposed to learn temporal correlations from sequential data by introducing recurrent connections in RBM. 
A structured RTRBM (sRTRBM)~\cite{mittelman2014structured} scaled up RTRBM by learning dependency structures between observations and hidden variables from data. 
More recently, Michalski et al.~\cite{michalski2014modeling} proposed a higher-order gated autoencoder that defines multiplicative interactions between consecutive frames and mapping units, and showed that temporal prediction problem can be viewed as learning and inferring higher-order interactions between consecutive images. 
Srivastava et al.~\cite{srivastava2015unsupervised} applied a sequence-to-sequence learning framework~\cite{sutskever2014sequence} to a video domain, and showed that long short-term memory (LSTM)~\cite{hochreiter1997long} networks are capable of generating video of bouncing handwritten digits.
In contrast to these previous studies, 
this paper tackles problems where control variables affect temporal dynamics, and in addition 
scales up spatio-temporal prediction to larger-size images.

\cutparagraphup
\paragraph{ALE: Combining Deep Learning and RL.}
Atari 2600 games provide challenging environments for RL because of high-dimensional visual observations, partial observability, and delayed rewards. Approaches that combine deep learning and RL have made significant advances~\cite{mnih2013playing,mnih2015human,guo2014deep}. Specifically, DQN~\cite{mnih2013playing} combined Q-learning~\cite{watkins1992q} with a convolutional neural network (CNN) and achieved state-of-the-art performance on many Atari games. Guo et al.~\cite{guo2014deep} used the ALE-emulator for making action-conditional predictions with slow UCT~\cite{kocsis2006bandit}, a Monte-Carlo tree search method, to generate training data for a fast-acting CNN, which outperformed DQN on several domains. Throughout this paper we will use DQN to refer to the architecture used in~\cite{mnih2013playing} (a more recent work~\cite{mnih2015human} used a deeper CNN with more data to produce the currently best-performing Atari game players). 

\cutparagraphup
\paragraph{Action-Conditional Predictive Model for RL.}
The idea of building a predictive model for vision-based RL problems was introduced by Schmidhuber and Huber~\cite{schmidhuber1991learning}. They proposed a neural network that predicts the attention region given the previous frame and an \textit{attention-guiding} action. More recently, Lenz et al.~\cite{lenz_deepMPC_2015} proposed a recurrent neural network with multiplicative interactions that predicts the physical coordinate of a robot. Compared to this previous work, our work is evaluated on much higher-dimensional data with complex dependencies among observations.
There have been a few attempts to learn from ALE data a transition-model that makes predictions of future frames. One line of work~\cite{bellemare2013bayesian,bellemare2014skip} divides game images into patches and applies a Bayesian framework to predict patch-based observations. However, this approach assumes that neighboring patches are enough to predict the center patch, which is not true in Atari games because of many complex interactions. 
The evaluation in this prior work is 1-step prediction loss; in contrast, here we make and evaluate long-term predictions both for quality of pixels generated and for usefulness to control.

\cutsectionup 
\section{Proposed Architectures and Training Method} \label{headings}
\cutsectiondown

\begin{figure}
    \centering
    \begin{subfigure}{0.47\textwidth}
	    \includegraphics[width=\linewidth]{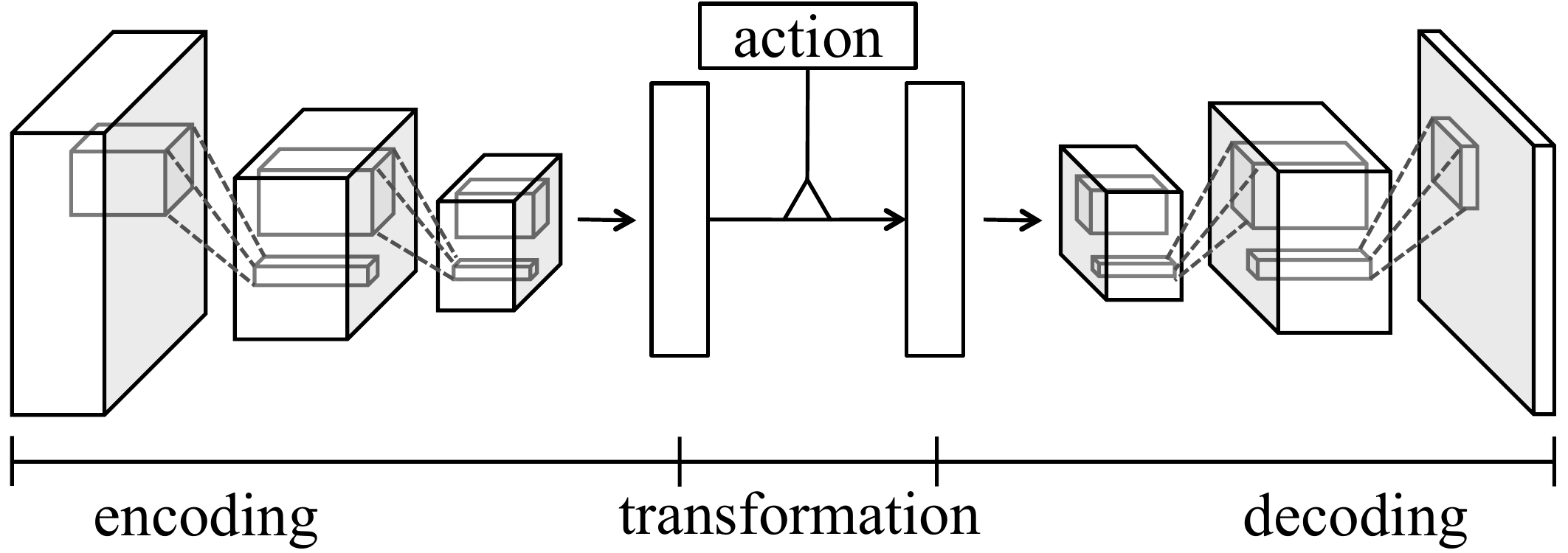} 
   	    \caption{\A{}}
   	    \label{fig:arch-cnn}
   	\end{subfigure}
   	\hspace{\fill}
	\begin{subfigure}{0.47\textwidth}
	    \includegraphics[width=\linewidth]{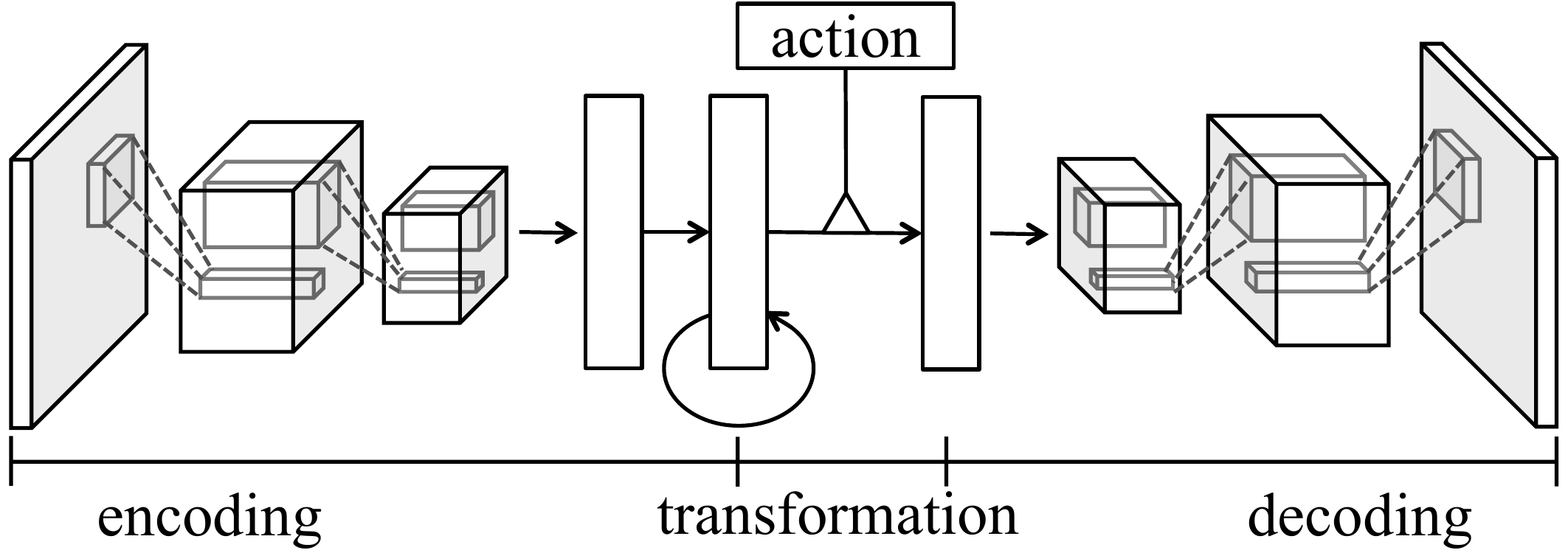} 
   	    \caption{\B{}}
   	    \label{fig:arch-lstm}
	\end{subfigure}
	\vspace{-5pt}
\caption{Proposed Encoding-Transformation-Decoding network architectures.} 
\label{fig:arch}
\vspace*{-0.2in}
\end{figure}

The goal of our architectures is to learn a function $f:\textbf{x}_{1:t},\textbf{a}_{t} \rightarrow \textbf{x}_{t+1}$,  where $\textbf{x}_{t}$ and $\textbf{a}_t$ are the frame and action variables at time $t$, and $\textbf{x}_{1:t}$ are the frames from time $1$ to time $t$. Figure~\ref{fig:arch} shows our two architectures that are each composed of encoding layers that extract spatio-temporal features from the input frames (\S\ref{sub:encoding}), action-conditional transformation layers that transform the encoded features into a prediction of the next frame in high-level feature space by introducing action variables as additional input (\S\ref{sub:transformation}) and finally decoding layers that map the predicted high-level features into pixels (\S\ref{sub:decoding}).
Our contributions are in the novel action-conditional deep convolutional architectures for high-dimensional, long-term prediction as well as in the novel use of the architectures in vision-based RL domains. 

\cutsubsectionup
\subsection{Two Variants: Feedforward Encoding and Recurrent Encoding} \label{sub:encoding}
\cutsubsectiondown


\textbf{\A{}} takes a fixed history of previous frames as an input, which is concatenated through channels (Figure \ref{fig:arch-cnn}), and stacked convolution layers extract spatio-temporal features directly from the concatenated frames.
The encoded feature vector $\mathbf{h}^{enc}_{t}\in\mathbb{R}^{n}$ at time $t$ is: 
\begin{equation}
\mathbf{h}^{enc}_{t} = \mbox{CNN}\left(\mathbf{x}_{t-m+1:t}\right),
\end{equation}
where $\mathbf{x}_{t-m+1:t}\in\mathbb{R}^{(m\times c)\times h\times w}$ denotes $m$ frames of $h \times w$ pixel images with $c$ color channels.
CNN is a mapping from raw pixels to a high-level feature vector using multiple convolution layers and a fully-connected layer at the end, each of which is followed by a non-linearity. 
This encoding can be viewed as \textit{early-fusion} \cite{karpathy2014large} (other types of fusions, e.g., \textit{late-fusion} or 3D convolution \cite{DBLP:journals/corr/TranBFTP14} can also be applied to this architecture). 

\textbf{\B{}} takes one frame as an input for each time-step and extracts spatio-temporal features using an RNN in which the temporal dynamics is modeled by the recurrent layer on top of the high-level feature vector extracted by convolution layers (Figure \ref{fig:arch-lstm}). In this paper, LSTM without peephole connection is used for the recurrent layer as follows:
\begin{equation}
\left[\mathbf{h}_{t}^{enc},\mathbf{c}_{t}\right]=\mbox{LSTM}\left(\mbox{CNN}\left(\mathbf{x}_{t}\right),\mathbf{h}_{t-1}^{enc},\mathbf{c}_{t-1}\right),
\end{equation}
where $\textbf{c}_{t}\in\mathbb{R}^{n}$ is a \textit{memory cell} that retains information from a deep history of inputs. Intuitively, $\mbox{CNN}\left(\mathbf{x}_{t}\right)$ is given as input to the LSTM so that the LSTM captures temporal correlations from high-level spatial features.

\cutsubsectionup
\subsection{Multiplicative Action-Conditional Transformation} \label{sub:transformation}
\cutsubsectiondown

We use multiplicative interactions between the encoded feature vector and the control variables:
\vspace*{-3pt}
\begin{equation}
h_{t,i}^{dec}=\sum_{j,l}W_{ijl}h_{t,j}^{enc}a_{t,l}+b_{i},
\vspace*{-3pt}
\end{equation}
where $\mathbf{h}^{enc}_t\in\mathbb{R}^n$ is an encoded feature, $\mathbf{h}^{dec}_t\in\mathbb{R}^n$ is an action-transformed feature, $\mathbf{a}_{t}\in\mathbb{R}^a$ is the action-vector at time $t$, $\mathbf{W}\in\mathbb{R}^{n \times n \times a}$ is 3-way tensor weight, and $\mathbf{b}\in\mathbb{R}^{n}$ is bias. 
When the action $\textbf{a}$ is represented using one-hot vector, using a 3-way tensor is equivalent to using different weight matrices for each action. This enables the architecture to model different transformations for different actions. The advantages of multiplicative interactions have been explored in image and text processing ~\cite{taylor2009factored, sutskever2011generating, memisevic2013learning}.
In practice the 3-way tensor is not scalable because of its large number of parameters.
Thus, we approximate the tensor by factorizing into three matrices as follows~\cite{taylor2009factored}:
\begin{equation} \label{eq:factorization}
\mathbf{h}^{dec}_t=\mathbf{W}^{dec}\left(\mathbf{W}^{enc}\mathbf{h}^{enc}_t\odot\mathbf{W}^{a}\mathbf{a}_t\right) + \mathbf{b},
\end{equation} 
where $\mathbf{W}^{dec}\in\mathbb{R}^{n \times f},\mathbf{W}^{enc}\in\mathbb{R}^{f \times n},\mathbf{W}^{a}\in\mathbb{R}^{f \times a},\mathbf{b}\in\mathbb{R}^n$, and $f$ is the number of factors. Unlike the 3-way tensor, the above factorization shares the weights between different actions by mapping them to the size-$f$ factors. 
This sharing may be desirable relative to the 3-way tensor when there are common temporal dynamics in the data across different actions (discussed further in \S\ref{sub:analysis}).

\cutsubsectionup
\subsection{Convolutional Decoding} \label{sub:decoding}
\cutsubsectiondown

It has been recently shown that a CNN is capable of generating an image effectively using upsampling followed by convolution with stride of 1~\cite{dosovitskiy2014learning}. Similarly, we use the ``inverse" operation of convolution, called deconvolution, which maps $1\times1$ spatial region of the input to $d\times d$ using deconvolution kernels. The effect of $s\times s$ upsampling can be achieved without explicitly upsampling the feature map by using stride of $s$. We found that this operation is more efficient than upsampling followed by convolution because of the smaller number of convolutions with larger stride.

In the proposed architecture, the transformed feature vector $\mathbf{h}^{dec}$ is decoded into pixels as follows:
\begin{equation}
\hat{\mathbf{x}}_{t+1} = \mbox{Deconv}\left(\mbox{Reshape}\left(\mathbf{h}^{dec}\right)\right),
\end{equation}
where Reshape is a fully-connected layer where hidden units form a 3D feature map, and Deconv consists of multiple deconvolution layers, each of which is followed by a non-linearity except for the last deconvolution layer.

\cutsubsectionup
\subsection{Curriculum Learning with Multi-Step Prediction} \label{sub:training}
\cutsubsectiondown

It is almost inevitable for a predictive model to make noisy predictions 
of high-dimensional images.
When the model is trained on a 1-step prediction objective, 
small prediction errors can compound 
through time.
To alleviate this effect, we use a multi-step prediction objective. 
More specifically, given the training data $D=\left\{ \left(\left(\mathbf{x}_{1}^{(i)},\mathbf{a}_{1}^{(i)}\right),...,\left(\mathbf{x}_{T_{i}}^{(i)},\mathbf{a}_{T_{i}}^{(i)}\right)\right)\right\} _{i=1}^{N}$, the model is trained to minimize the average squared error over $K$-step predictions as follows:
\begin{equation}
\mathcal{L}_{K}\left(\theta\right)=\frac{1}{2K}\sum_{i}\sum_{t}\sum_{k=1}^{K}\left\Vert \hat{\mathbf{x}}_{t+k}^{(i)}-\mathbf{x}_{t+k}^{(i)}\right\Vert ^{2},
\end{equation}
where $\hat{\mathbf{x}}_{t+k}^{(i)}$ is a $k$-step future prediction. 
Intuitively, the network is repeatedly \textit{unrolled} through $K$ time steps by using its prediction as an input for the next time-step. 

The model is trained in multiple phases based on increasing $K$ as suggested by Michalski et al.~\cite{michalski2014modeling}. 
In other words, the model is trained to predict short-term future frames and fine-tuned to predict longer-term future frames after the previous phase converges. 
We found that this curriculum learning~\cite{bengio2009curriculum} approach is necessary to stabilize the training.
A stochastic gradient descent with backpropagation through time (BPTT) is used to optimize the parameters of the network.

\cutsectionup 
\section{Experiments} \label{experiment}
\cutsectiondown

In the experiments that follow, we have the following goals for our two architectures. 1) To evaluate the predicted frames in two ways: qualitatively evaluating the generated video, and quantitatively evaluating the pixel-based squared error, 2) To evaluate the usefulness of predicted frames for control in two ways: by replacing the emulator's frames with predicted frames for use by DQN, and by using the predictions to improve exploration in DQN, and 3) To analyze the representations learned by our architectures.
We begin by describing the details of the data, and model architecture, and baselines. 

\cutparagraphup
\paragraph{Data and Preprocessing.}
We 
used our replication of DQN to generate game-play video datasets using an $\epsilon$-greedy policy with $\epsilon=0.3$, i.e.\ DQN is forced to choose a random action with 30$\%$ probability.
For each game, the dataset consists of about $500,000$ training frames and $50,000$ test frames with actions chosen by DQN. 
Following DQN, actions are chosen once every $4$ frames which reduces the video from 60fps to 15fps. 
The number of actions available in games varies from $3$ to $18$, and they are represented as one-hot vectors.
We used full-resolution RGB images ($210\times160$) and preprocessed the images by subtracting mean pixel values and dividing each pixel value by $255$.

\cutparagraphup
\paragraph{Network Architecture.}
Across all game domains, we use the same network architecture as follows.
The encoding layers consist of $4$ convolution layers and one fully-connected layer with $2048$ hidden units. 
The convolution layers use $64$ $(8\times8)$, $128$ $(6\times6)$, $128$ $(6\times6)$, and $128$ $(4\times4)$ filters with stride of 2. Every layer is followed by a rectified linear function~\cite{nair2010rectified}. 
In the \BBB{} network, an LSTM layer with $2048$ hidden units is added on top of the fully-connected layer.
The number of factors in the transformation layer is $2048$. 
The decoding layers consists of one fully-connected layer with $11264$ $(=128\times11\times8)$ hidden units followed by $4$ deconvolution layers.
The deconvolution layers use $128$ $(4\times4)$, $128$ $(6\times6)$, $128$ $(6\times6)$, and $3$ $(8\times8)$ filters with stride of 2.
For the \AAA{} network, the last $4$ frames are given as an input for each time-step. 
The \BBB{} network takes one frame for each time-step, but it is unrolled through the last $11$ frames to initialize the LSTM hidden units before making a prediction. Our implementation is based on Caffe toolbox \cite{jia2014caffe}.

\cutparagraphup
\paragraph{Details of Training.}
We use the curriculum learning scheme above with three phases of increasing prediction step objectives of $1$, $3$ and $5$ steps, and learning rates of $10^{-4}$, $10^{-5}$, and $10^{-5}$, respectively. 
RMSProp~\cite{tieleman2012, graves2013generating} is used with momentum of $0.9$, (squared) gradient momentum of $0.95$, and min squared gradient of $0.01$. 
The batch size for each training phase is $32$, $8$, and $8$ for the \AAA{} network and $4$, $4$, and $4$ for the \BBB{} network, respectively.
When the \BBB{} network is trained on 1-step prediction objective, the network is unrolled through $20$ steps and predicts the last $10$ frames by taking ground-truth images as input. Gradients are clipped at $[-0.1, 0.1]$ before non-linearity of each gate of LSTM as suggested by~\cite{graves2013generating}.

\cutparagraphup
\paragraph{Two Baselines for Comparison.}
The first baseline is a multi-layer perceptron (\textit{MLP}) that takes the last frame as input and has 4 hidden layers with 400, 2048, 2048, and 400 units. The action input is concatenated to the second hidden layer. 
This baseline uses approximately the same number of parameters as the  \BBB{} model. 
The second baseline, no-action feedforward (or \textit{naFf}), is the same as the \AAA{} model (Figure~\ref{fig:arch-cnn}) except that the transformation layer consists of one fully-connected layer that does not get the action as input. 

\begin{figure}
    \centering
   \includegraphics[width=0.8\textwidth]{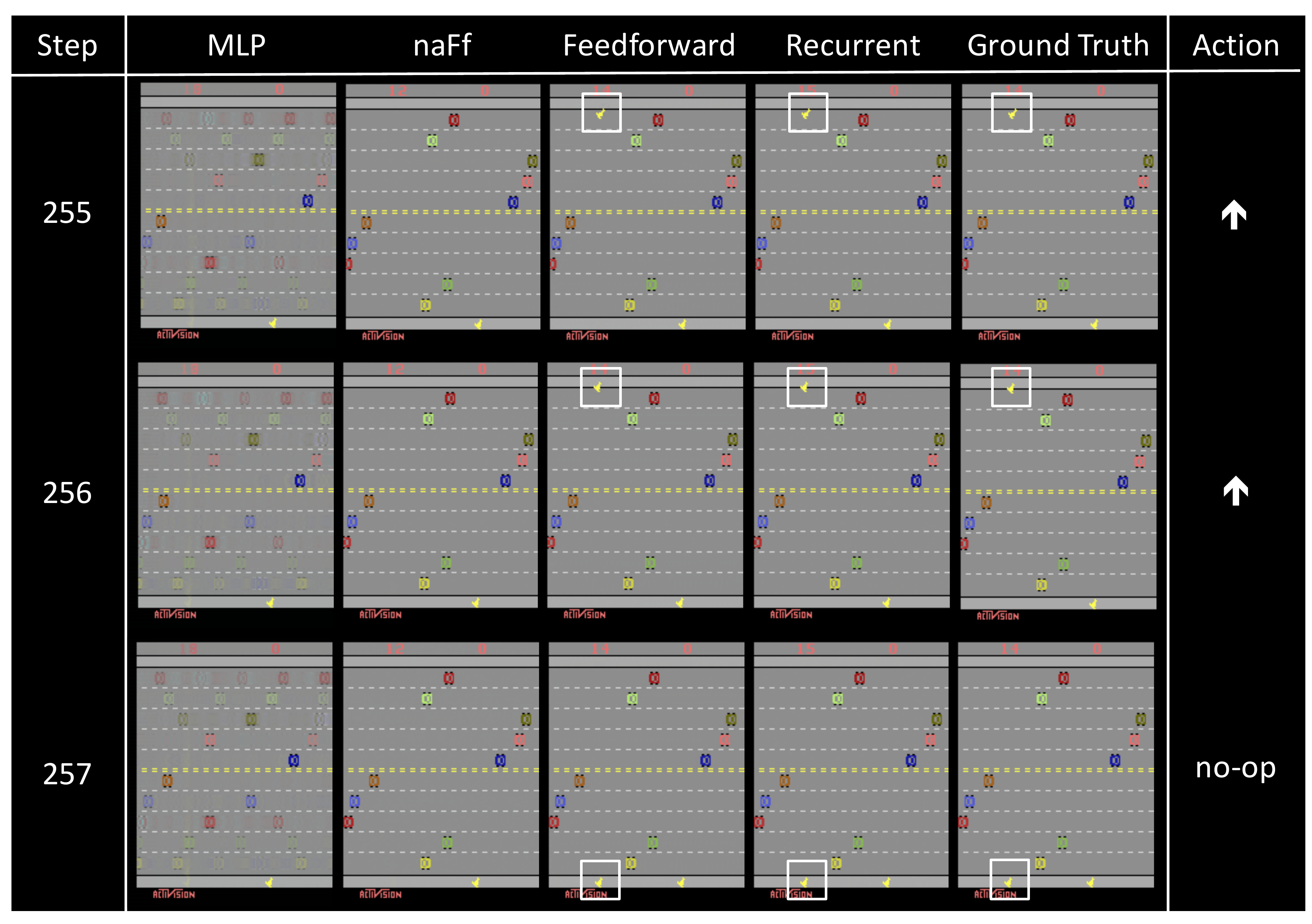}
    \vspace{-5pt}
	\caption{Example of predictions over 250 steps in Freeway. The `Step' and `Action' columns show the number of prediction steps and the actions taken respectively. The white boxes indicate the object controlled by the agent. From prediction step 256 to 257 the controlled object crosses the top boundary and reappears at the bottom; this non-linear shift is predicted by our architectures and is not predicted by MLP and naFf. The horizontal movements of the uncontrolled objects are predicted by our architectures and naFf but not by MLP.}
    \label{fig:video}
    \vspace{-20pt}
\end{figure}

\cutsubsectionup 
\subsection{Evaluation of Predicted Frames} \label{sub:exp-qual} 
\cutsubsectiondown

\cutparagraphup
\paragraph{Qualitative Evaluation: Prediction video.}
The prediction videos of our models and baselines are available in the supplementary material and at the following website: {\small \url{https://sites.google.com/a/umich.edu/junhyuk-oh/action-conditional-video-prediction}}. As seen in the videos, the proposed models make qualitatively reasonable predictions over $30$--$500$ steps depending on the game.
In all games, the MLP baseline quickly diverges, and the naFf baseline fails to predict the controlled object.
An example of long-term predictions is illustrated in Figure~\ref{fig:video}. 
We observed that both of our models predict complex local translations well such as the movement of vehicles and the controlled object.
They can predict interactions between objects such as collision of two objects.
Since our architectures effectively extract hierarchical features using CNN, they are able to make a prediction that requires a global context. 
For example, in Figure~\ref{fig:video}, the model predicts the sudden change of the location of the controlled object (from the top to the bottom) at 257-step.

However, both of our models have difficulty in accurately predicting small objects, such as bullets in Space Invaders. 
The reason is that the squared error signal is small when the model fails to predict small objects during training.
Another difficulty is in handling stochasticity. 
In Seaquest, e.g., new objects appear from the left side or right side randomly, and so are hard to predict.
Although our models do generate new objects with reasonable shapes and movements (e.g., after appearing they move as in the true frames), the generated frames do not necessarily match the ground-truth. 

\begin{figure}
  \centering
  \begin{subfigure}{0.17\textwidth}
  	  \centering
	  \includegraphics[width=\linewidth]{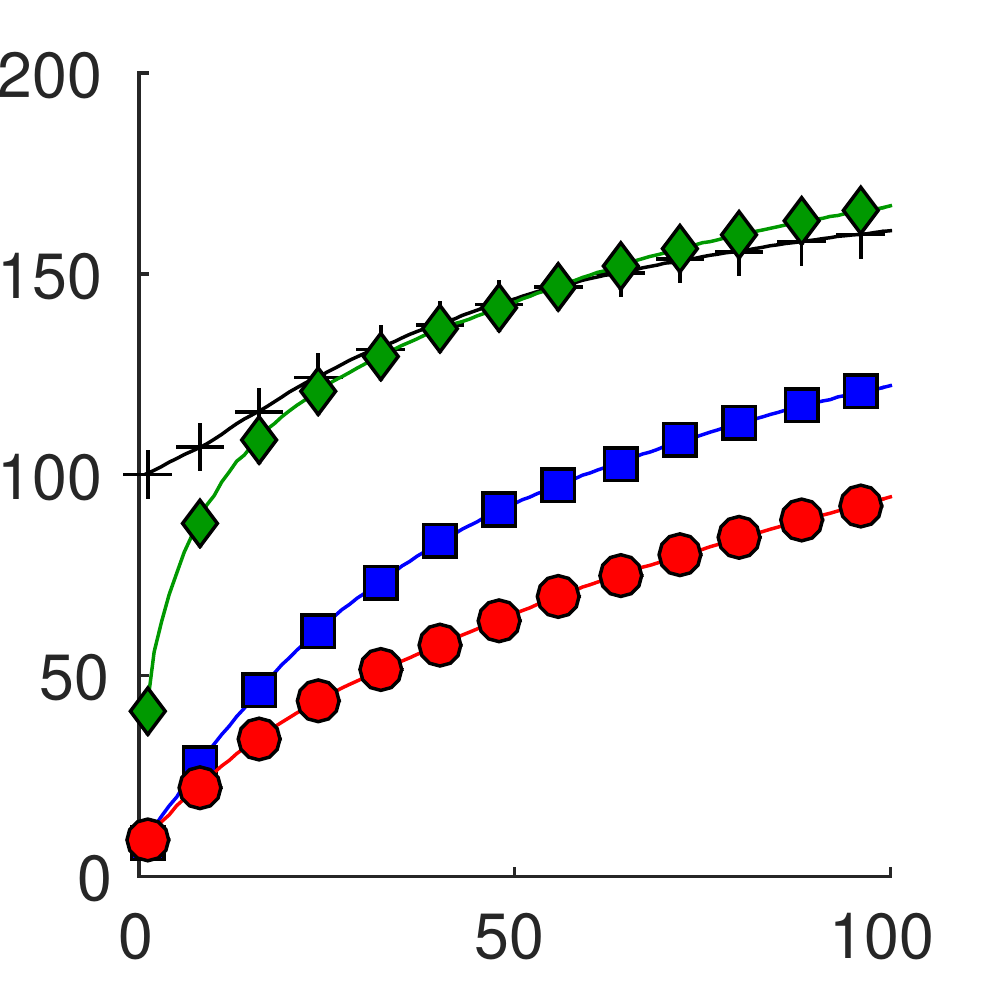}
	  \caption{Seaquest}
  \end{subfigure}
  \begin{subfigure}{0.17\textwidth}
	  \centering
	  \includegraphics[width=\linewidth]{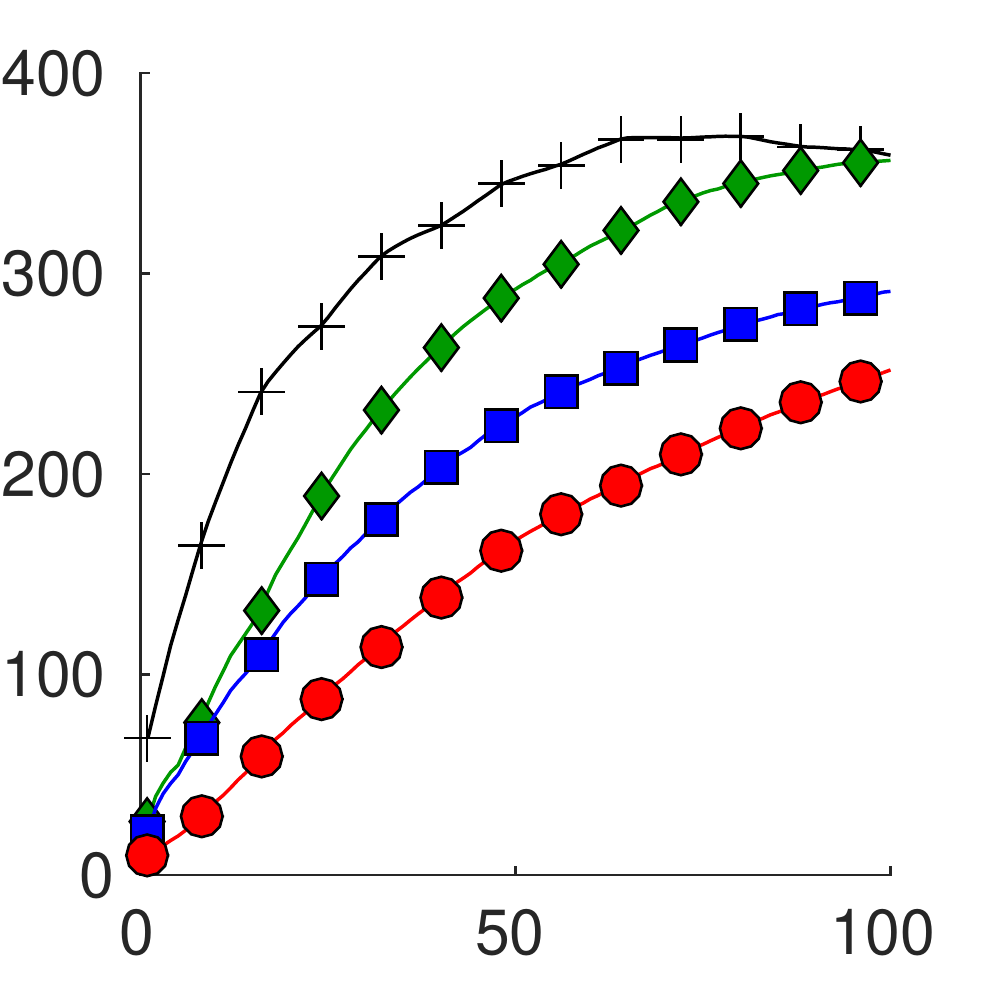}
  	  \caption{Space Invaders}
  \end{subfigure}
  \begin{subfigure}{0.17\textwidth}
	  \centering	
	  \includegraphics[width=\linewidth]{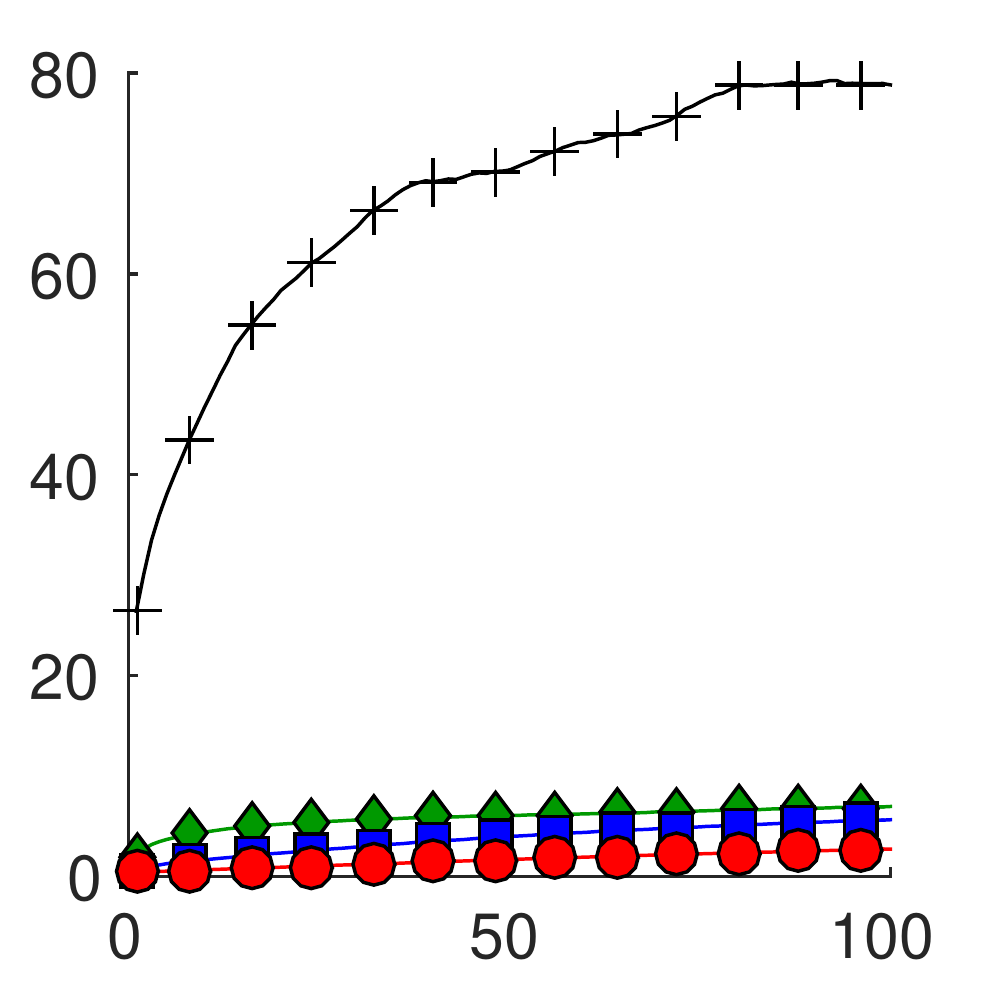}
  	  \caption{Freeway}
  \end{subfigure}
  \begin{subfigure}{0.17\textwidth}
	  \centering	
	  \includegraphics[width=\linewidth]{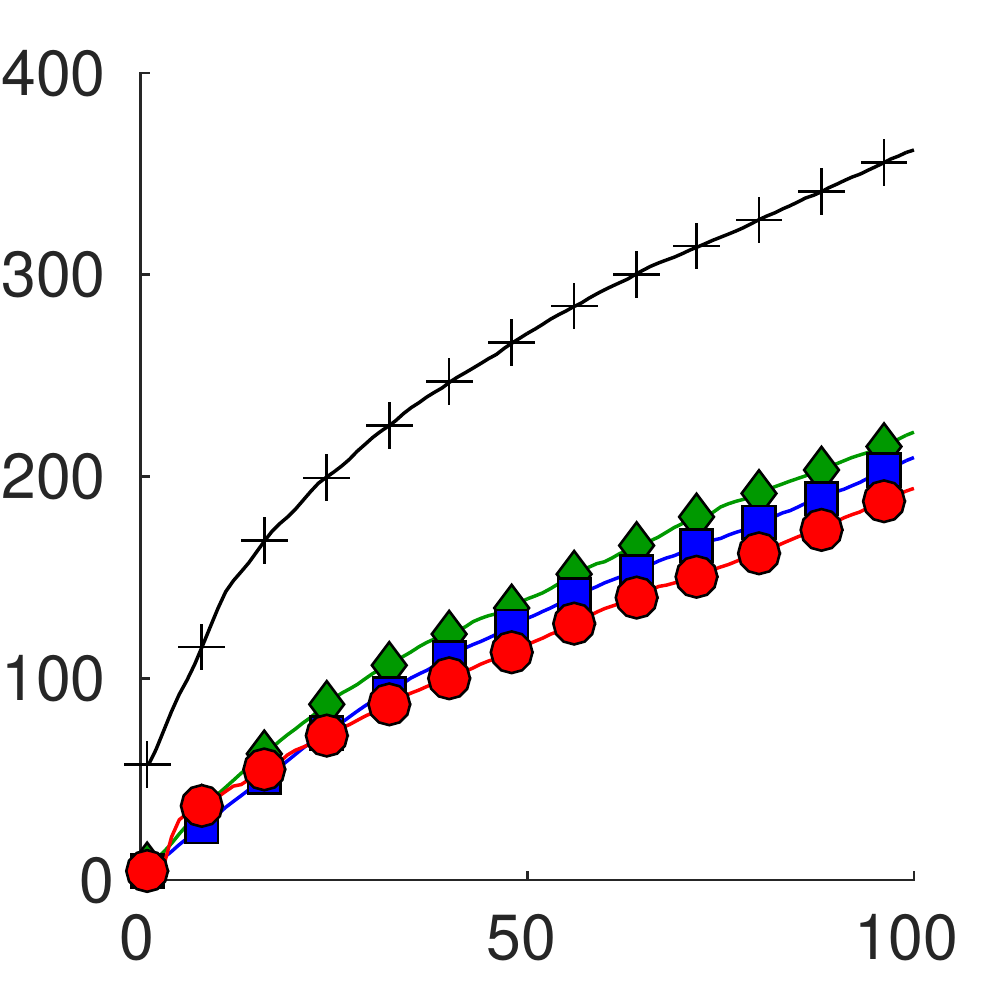}
  	  \caption{QBert}
  \end{subfigure}
  \begin{subfigure}{0.17\textwidth}
	  \centering	
	  \includegraphics[width=\linewidth]{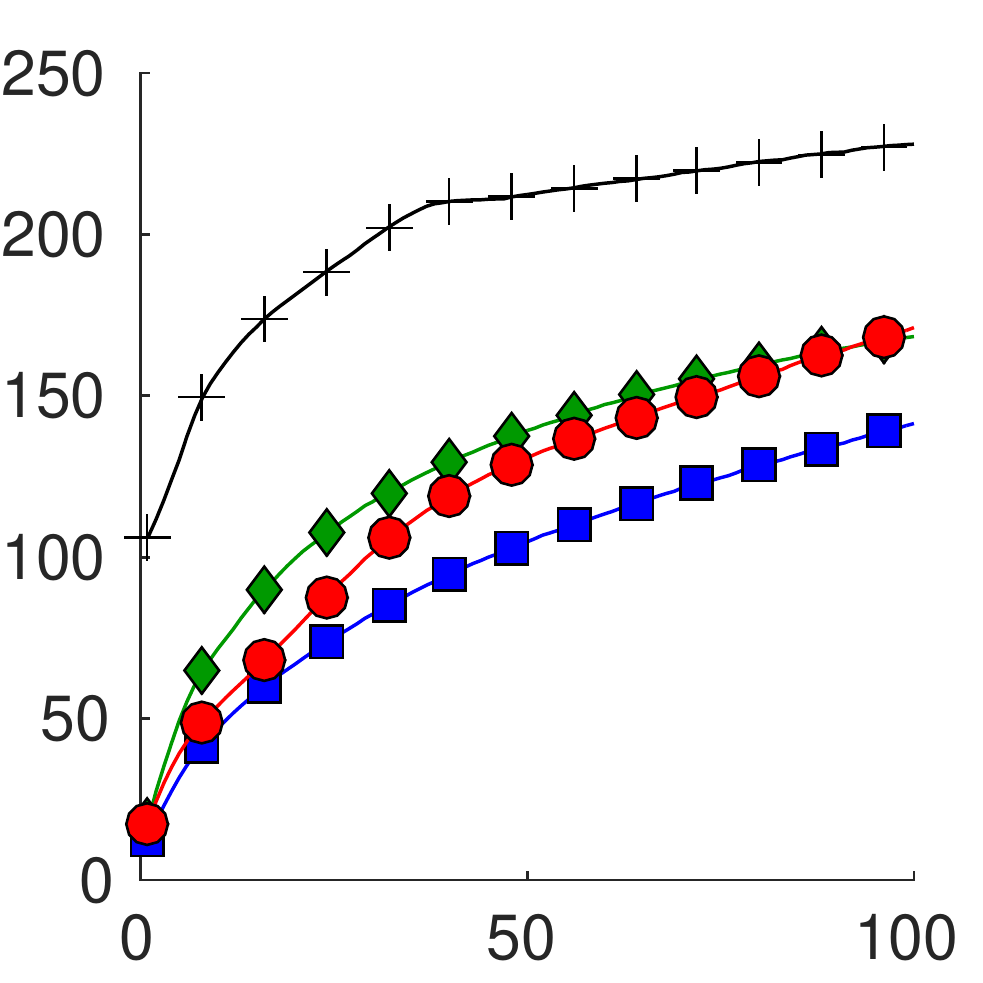}
	  \caption{Ms Pacman}
  \end{subfigure}
  \begin{subfigure}{0.12\textwidth}
	  \includegraphics[width=\linewidth]{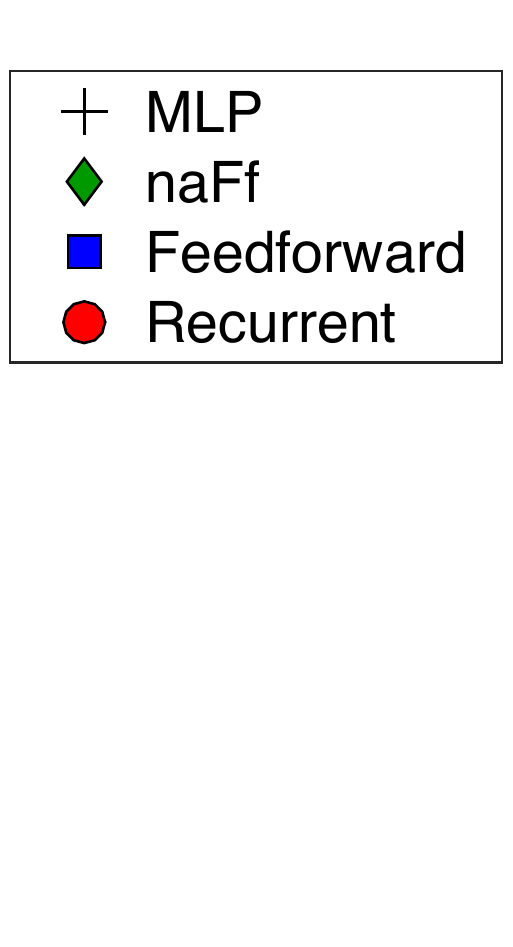}
  \end{subfigure}
  
\vspace*{-1.5ex}
  \caption{Mean squared error over 100-step predictions}
  \label{fig:plot-loss}
\vspace*{-20pt}  
\end{figure}

\cutparagraphup
\paragraph{Quantitative Evaluation: Squared Prediction Error.}
Mean squared error over 100-step predictions is reported in Figure~\ref{fig:plot-loss}. 
Our predictive models outperform the two baselines for all domains.
However, the gap between our predictive models and naFf baseline is not large except for Seaquest.
This is due to the fact that the object controlled by the action occupies only a small part of the image.

\begin{figure}
  \centering
  \small
  \begin{subfigure}{0.49\textwidth}
	  \centering
	  \includegraphics[width=0.99\linewidth]{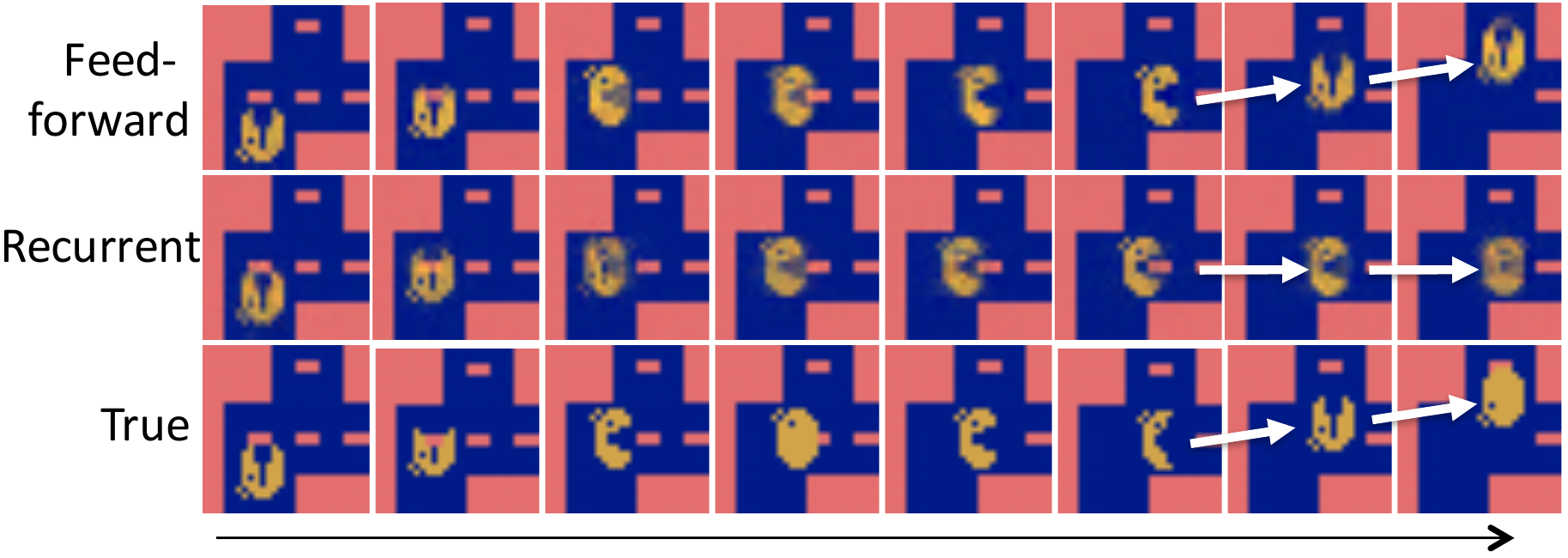}
	  \caption{Ms Pacman ($28 \times 28$ cropped)}
	  \label{fig:compare-pac}
  \end{subfigure}
  \begin{subfigure}{0.49\textwidth}
	  \centering
	  \includegraphics[width=0.99\linewidth]{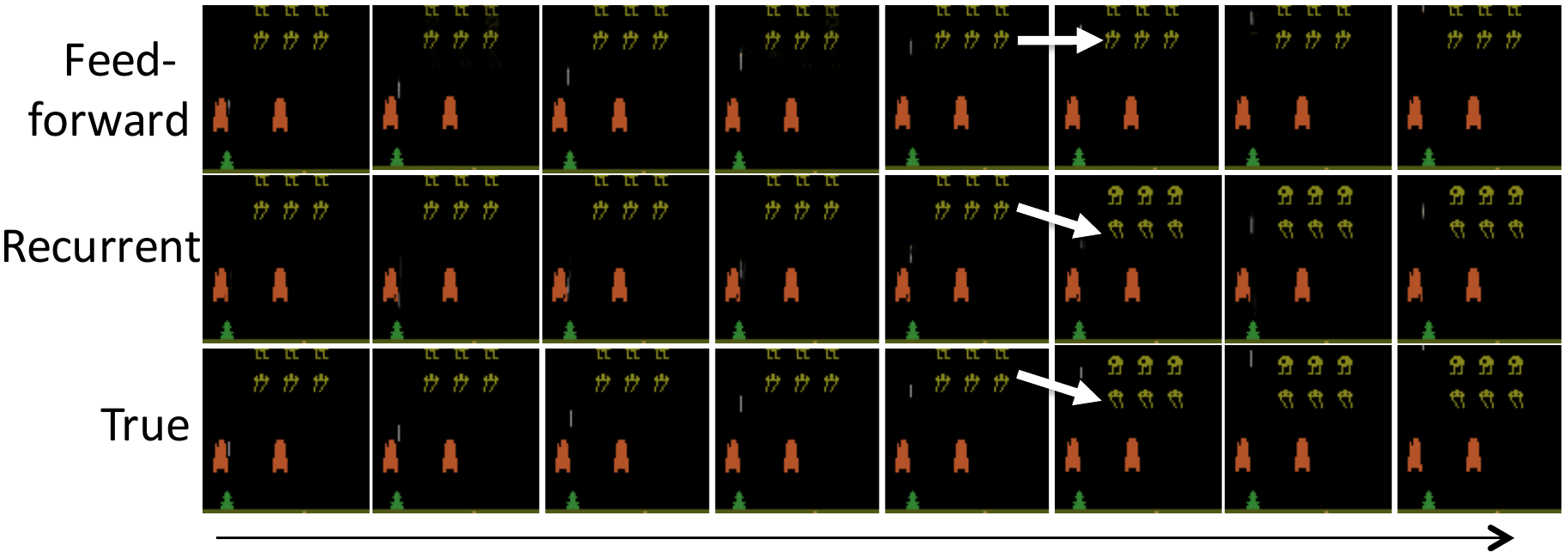}
	  \caption{Space Invaders ($90 \times 90$ cropped)}
	  \label{fig:compare-sp}
  \end{subfigure}
  
\vspace*{-5pt}
  
\caption{Comparison between two encoding models (feedforward and recurrent). (a) Controlled object is moving along a horizontal corridor. As the \BBB{} model makes a small translation error at 4th frame, 
the true position of the object is in the crossroad while the predicted position is still in the  corridor. The (true) object then moves upward which is not possible in the predicted position and so the predicted object keeps moving right. This is less likely to happen in \AAA{} because its position prediction is more accurate. (b) The objects move down after staying at the same location for the first five steps. The \AAA{} model fails to predict this movement because it only gets the last four frames as input, while the recurrent model predicts this downwards movement more correctly.
}
\label{fig:compare}
\end{figure}

\cutparagraphup
\paragraph{Qualitative Analysis of Relative Strengths and Weaknesses of Feedforward and Recurrent Encoding.}
\label{sub:comparison}
We hypothesize that \AAA{} can model more precise spatial transformations because its convolutional filters can learn temporal correlations directly from pixels in the concatenated frames. 
In contrast, convolutional filters in \BBB{} can learn only spatial features from the one-frame input, and the temporal context has to be captured by the recurrent layer on top of the high-level CNN features without localized information. 
On the other hand, \BBB{} is potentially better for modeling arbitrarily long-term dependencies, whereas \AAA{} is not suitable for long-term dependencies because it requires more memory and parameters as more frames are concatenated into the input. 

As evidence, in Figure~\ref{fig:compare-pac} we show a case where \AAA{} is better at predicting the precise movement of the controlled object, while \BBB{} makes a 1-2 pixel translation error. This small error leads to entirely different predicted frames after a few steps. Since the feedforward and recurrent architectures
are identical except for the encoding part, we conjecture that this result is due to the failure of precise spatio-temporal encoding in \BBB{}.
On the other hand, \BBB{} is better at predicting when the enemies move in Space Invaders (Figure~\ref{fig:compare-sp}). 
This is due to the fact that the enemies move after 9 steps, which is hard for \AAA{} to predict because it takes only the last four frames as input. 
We observed similar results showing that \AAA{} cannot handle long-term dependencies in other games. 

\cutsubsectionup
\subsection{Evaluating the Usefulness of Predictions for Control}
\begin{figure}
\vspace*{-10pt}
  \centering
  \begin{subfigure}{0.17\textwidth}
  	  \centering
	  \includegraphics[width=\linewidth]{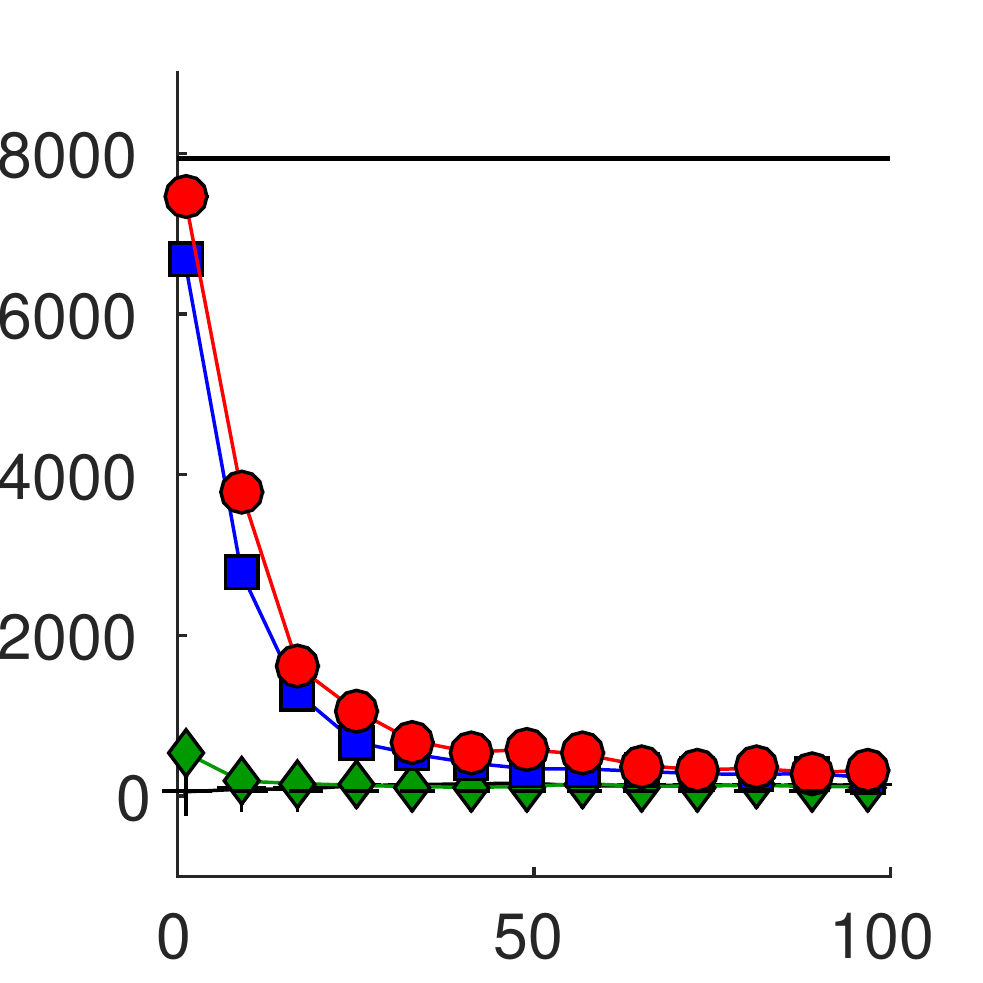}
	  \caption{Seaquest}
  \end{subfigure}
  \begin{subfigure}{0.17\textwidth}
	  \centering
	  \includegraphics[width=\linewidth]{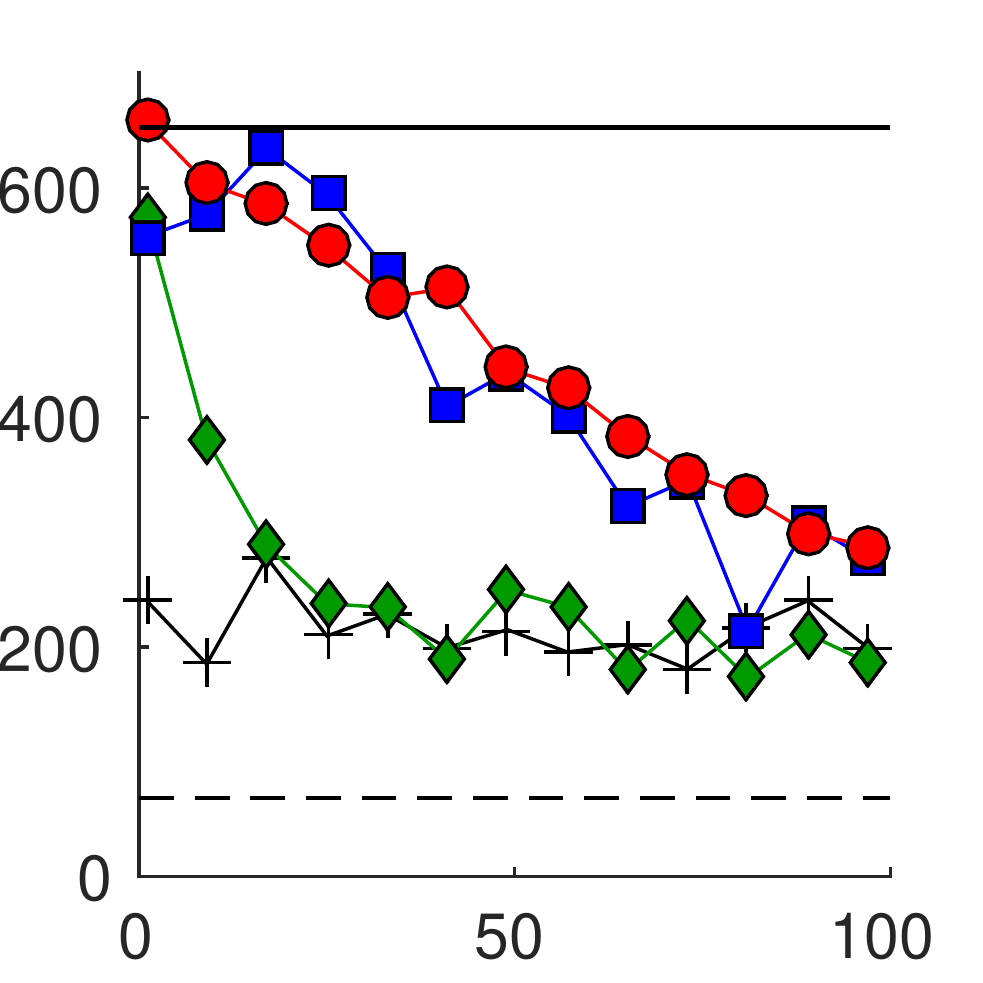}
  	  \caption{Space Invaders}
  \end{subfigure}
  \begin{subfigure}{0.17\textwidth}
	  \centering	
	  \includegraphics[width=\linewidth]{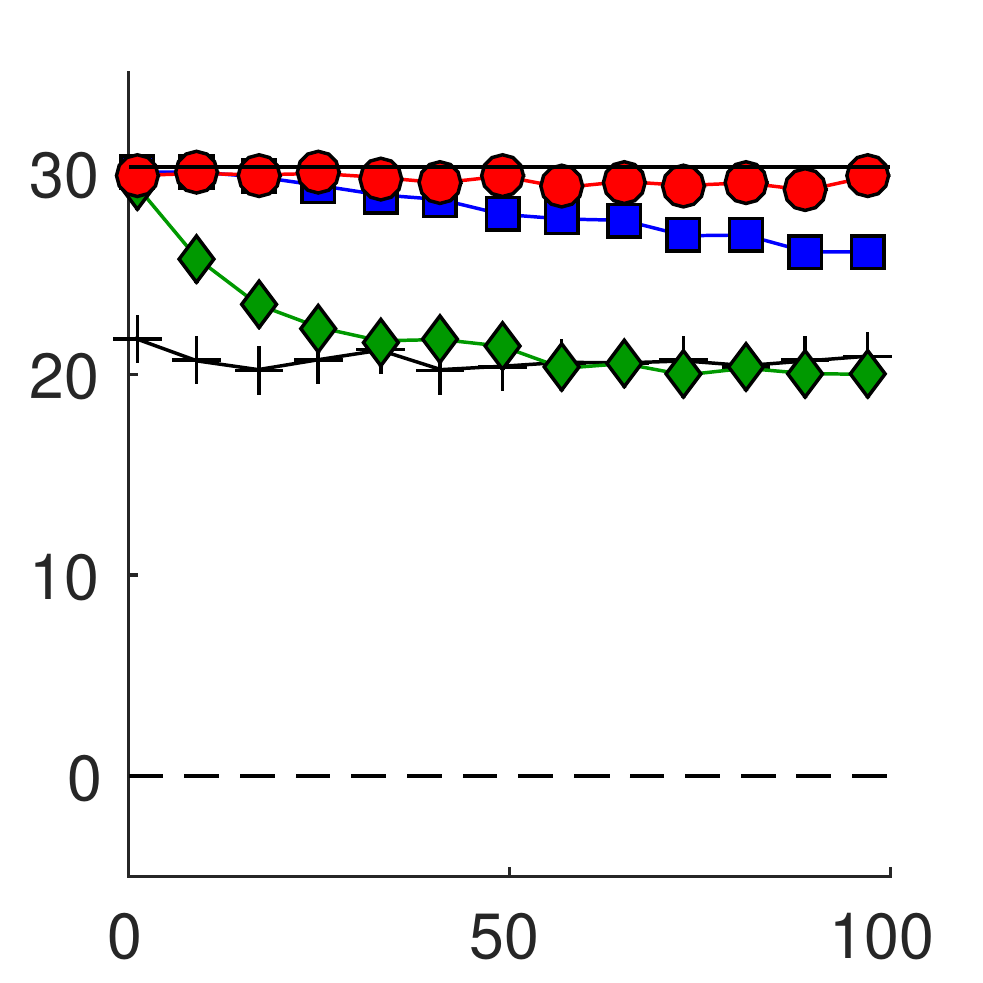}
  	  \caption{Freeway}
  \end{subfigure}
  \begin{subfigure}{0.17\textwidth}
	  \centering	
	  \includegraphics[width=\linewidth]{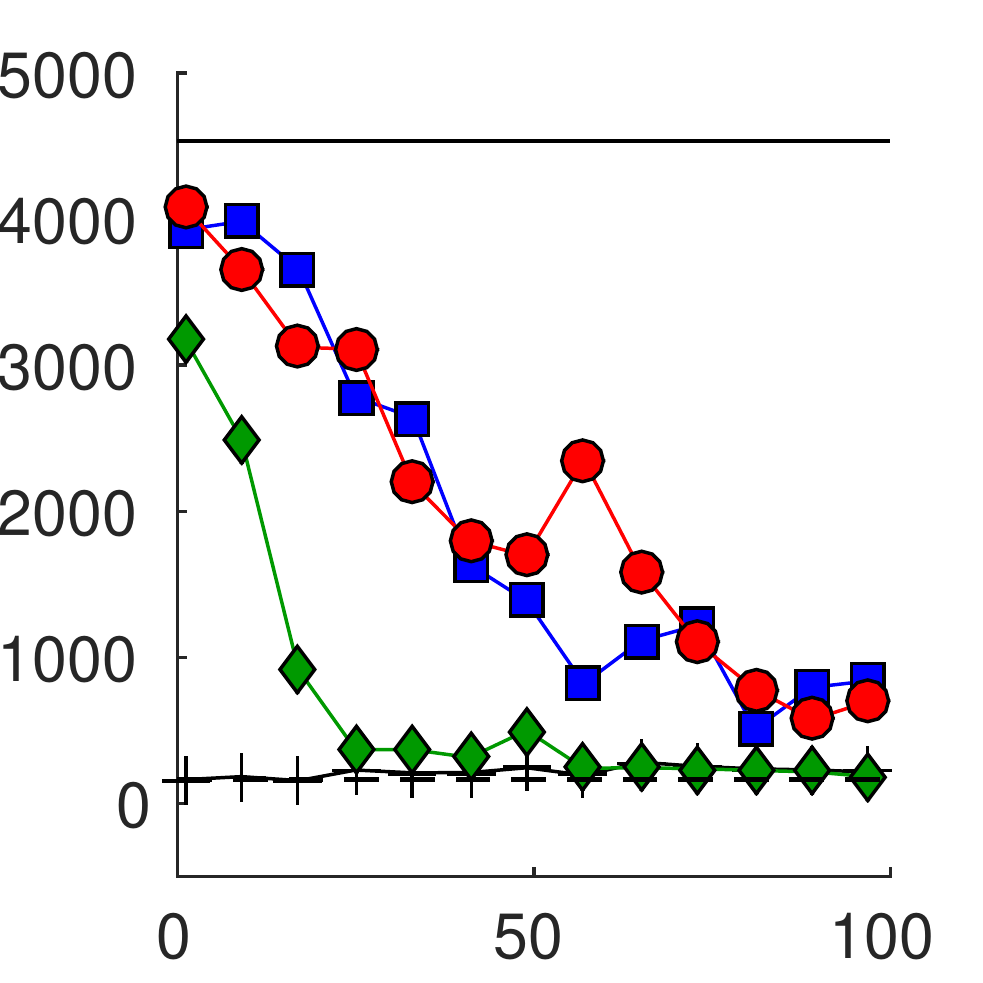}
  	  \caption{QBert}
  \end{subfigure}
  \begin{subfigure}{0.17\textwidth}
	  \centering	
	  \includegraphics[width=\linewidth]{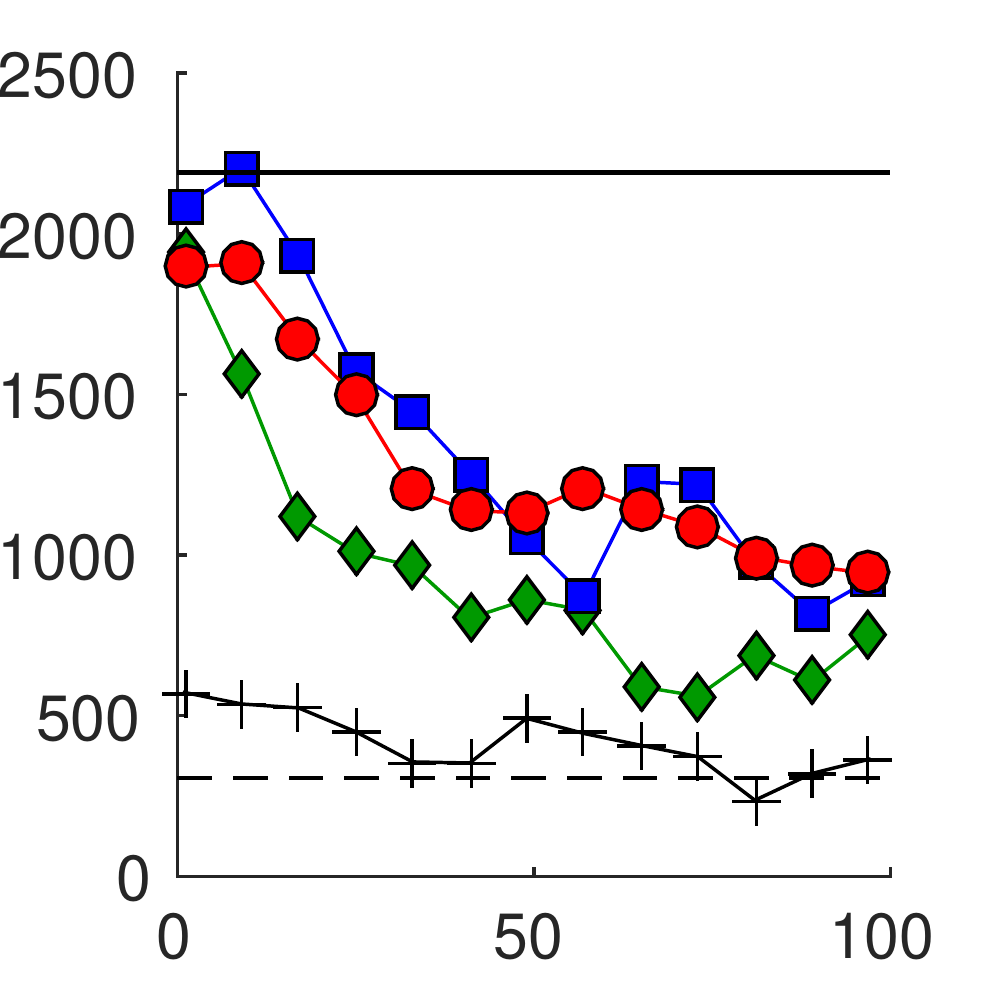}
	  \caption{Ms Pacman}
  \end{subfigure}
  \begin{subfigure}{0.12\textwidth}
	  \includegraphics[width=\linewidth]{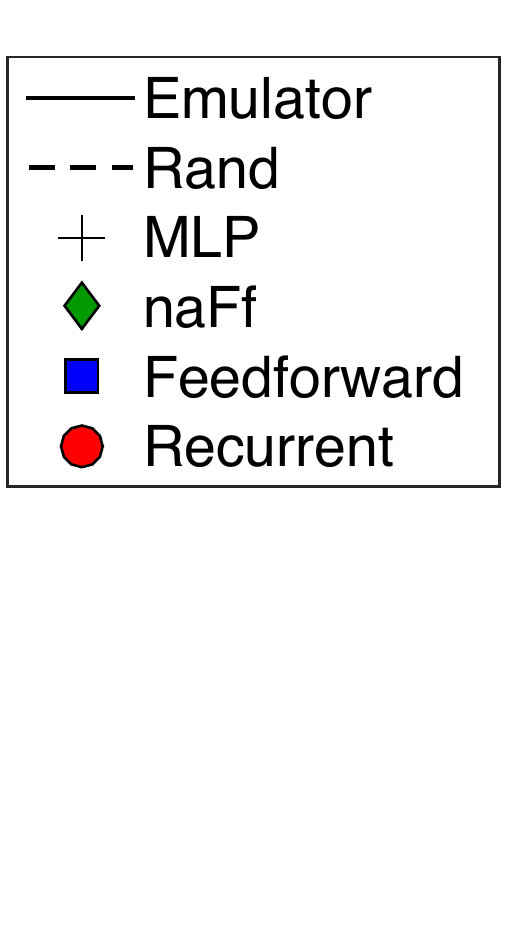}
  \end{subfigure}

\vspace{-5pt}
  \caption{Game play performance using the predictive model as an emulator. `Emulator' and `Rand' correspond to  the performance of DQN with true frames and random play respectively. The x-axis is the number of steps of prediction before re-initialization. The y-axis is the average game score measured from 30 plays.}
\label{fig:plot-tp-eval}
\vspace{-10pt}
\end{figure}

\cutparagraphup
\paragraph{Replacing Real Frames with Predicted Frames as Input to DQN.}
To evaluate how useful the predictions are for playing the games, we implement an 
evaluation method that uses the predictive model to replace the game emulator. More specifically, a DQN controller that takes the last four frames is first pre-trained using real frames and then used to play the games based on $\epsilon=0.05$-greedy policy where the input frames are generated by our predictive model instead of the game emulator. 
To evaluate how the depth of predictions influence the quality of control, we re-initialize the predictions using the true last frames after every n-steps of prediction for $1\leq n \leq 100$. Note that the DQN controller never takes a true frame, just the outputs of our predictive models.

The results are shown in Figure~\ref{fig:plot-tp-eval}. 
Unsurprisingly, replacing real frames with predicted frames reduces the score. However, in all the games using the model to repeatedly predict only a few time steps yields a score very close to that of using real frames. Our two architectures produce much better scores than the two baselines for deep predictions than would be suggested based on the much smaller differences in squared error. The likely cause of this is that our models are better able to predict the movement of the controlled object relative to the baselines even though such an ability may not always lead to better squared error. In three out of the five games the score remains much better than the score of random play even when using 100 steps of prediction.

\begin{table}[t]
\caption{\small{Average game score of DQN over 100 plays with standard error. The first row and the second row show the performance of our DQN replication with different exploration strategies.
}} 

\vspace{-5pt}

\centering
\small
\setlength{\tabcolsep}{4pt}
\begin{tabular}{lccccc}
\toprule
Model & Seaquest & S. Invaders & Freeway & QBert & Ms Pacman \\ 
\midrule
DQN - Random exploration & 13119 (538) & 698 (20) & 30.9 (0.2) & 3876 (106) & 2281 (53) \\ 
DQN - Informed exploration & 13265 (577) & 681 (23) & 32.2 (0.2) & 8238 (498) & 2522 (57) \\ 
\bottomrule
\end{tabular}
\label{table:game-score-partial}
\end{table}

\paragraph{Improving DQN via Informed Exploration.} 
To learn control in an RL domain, exploration of actions and states is necessary because without it the agent can get stuck in a bad sub-optimal policy. 
In DQN, the CNN-based agent was trained using an $\epsilon$-greedy policy in which the agent chooses either a greedy action or a random action by flipping a coin with probability of $\epsilon$.
Such random exploration is a basic strategy that produces sufficient exploration, but can be slower than more informed exploration strategies. 
Thus, we propose an \textit{informed exploration} strategy that follows the $\epsilon$-greedy policy, but chooses exploratory actions that lead to a frame that has been visited least often (in the last $d$ time steps), rather than random actions. Implementing this  strategy requires a predictive model because the next frame for each possible action has to be considered.

The method works as follows. The most recent $d$ frames are stored in a \textit{trajectory memory}, denoted $D=\left\{\mathbf{x}^{(i)}\right\}_{i=1}^{d}$. The predictive model is used to get the next frame ${\textbf{x}}^{(a)}$ for every action $a$. 
We estimate the visit-frequency for every predicted frame by summing the similarity between the predicted frame and the most $d$ recent frames stored in the trajectory memory using a Gaussian kernel as follows:%
\begin{equation}
n_{D}(\mathbf{x}^{(a)})=\sum_{i=1}^{d}k(\mathbf{x}^{(a)}, \mathbf{x}^{(i)});\quad 
k(\mathbf{x},\mathbf{y} )=\exp(-\sum_{j}\min(\max((x_{j}-y_{j})^{2}-\delta, 0),1)/\sigma)
\end{equation}
where $\delta$ is a threshold, and $\sigma$ is a kernel bandwidth.
The trajectory memory size is 200 for QBert and 20 for the other games, $\delta=0$ for Freeway and 50 for the others, and $\sigma=100$ for all games.
For computational efficiency, we trained a new \AAA{} network on $84\times84$ gray-scaled images as they are used as input for DQN. The details of the network architecture are provided in the supplementary material. 
Table~\ref{table:game-score-partial} summarizes the results.
The informed exploration improves DQN's performance using our predictive model in three of five games, with the most significant improvement in QBert.  Figure~\ref{fig:heatmap} shows how the informed exploration strategy improves the initial experience of DQN. 

\subsection{Analysis of Learned Representations} \label{sub:analysis}

\begin{figure}
    \begin{minipage}{0.65\linewidth}
        \begin{subfigure}{0.47\linewidth}
	        \centering
	        \includegraphics[width=\linewidth]{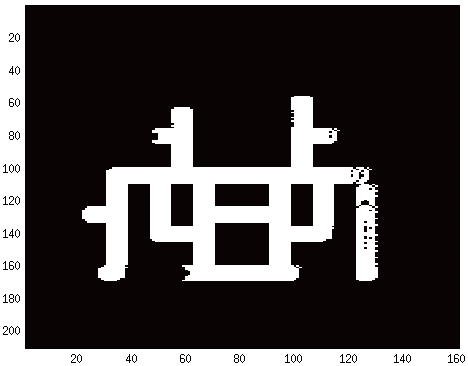}
	        \caption{Random exploration.}
        \end{subfigure}
        \begin{subfigure}{0.47\linewidth}
	        \centering	
	        \includegraphics[width=\linewidth]{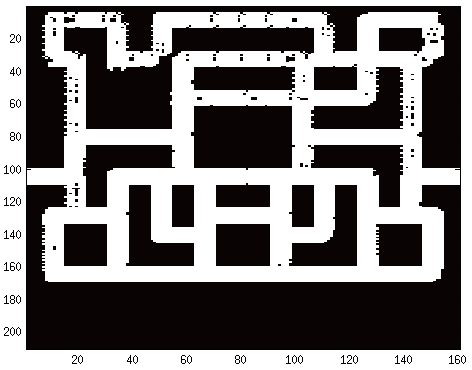}
	        \caption{Informed exploration.}
        \end{subfigure}
        \vspace{-5pt}
        \caption{Comparison between two exploration methods on Ms Pacman. Each heat map shows the trajectories of the controlled object measured over 2500 steps for the corresponding method. }
        \label{fig:heatmap}
    \end{minipage}
    \hfill
    \begin{minipage}{0.30\linewidth}
      \begin{center}
        \includegraphics[width=0.95\linewidth]{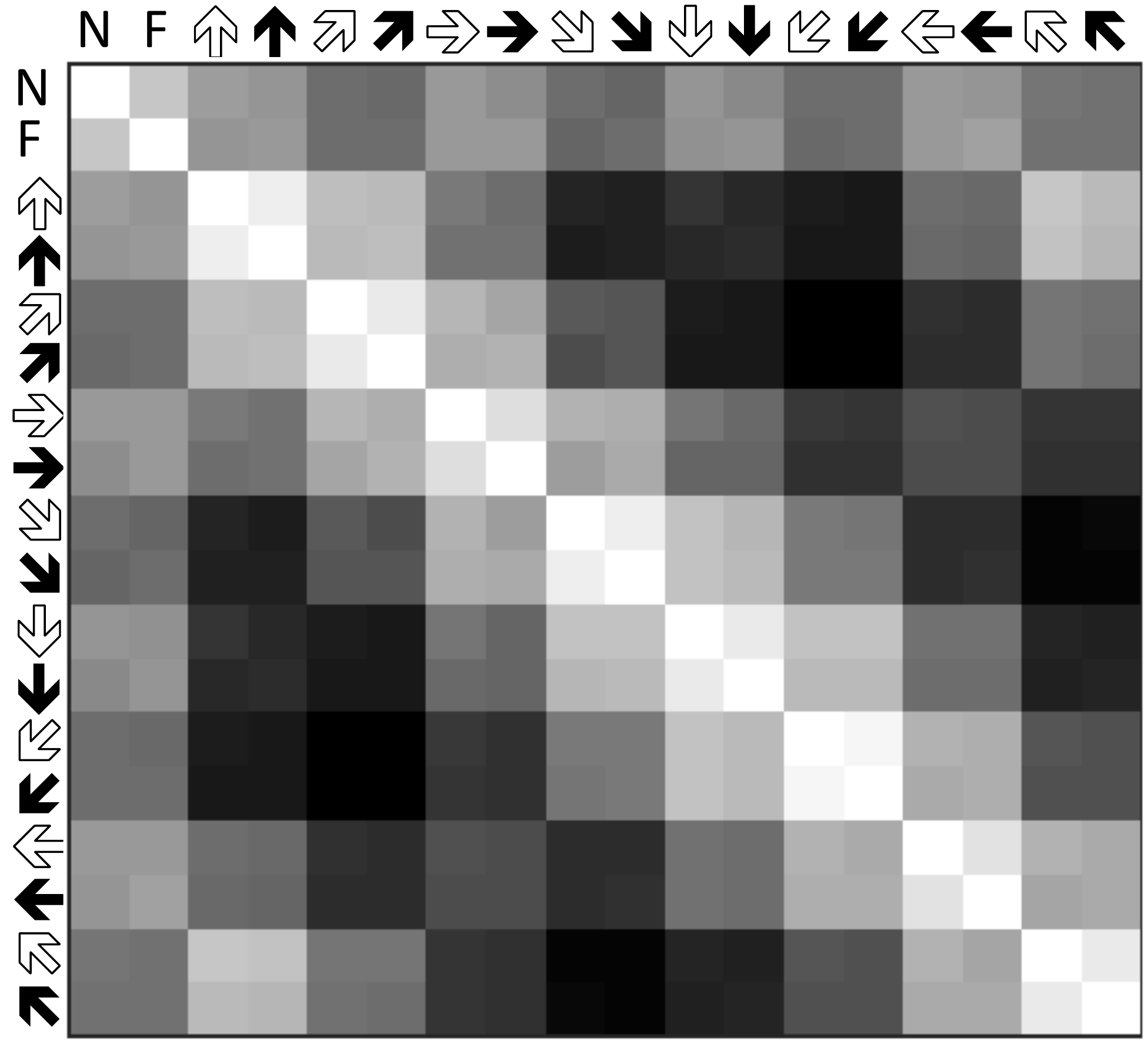}
      \end{center}
      \vspace{-5pt}
      \caption{Cosine similarity between every pair of action factors (see text for details).}
      \label{fig:action-correlation}
    \end{minipage}
    
    \vspace*{-10pt}
\end{figure}

\cutparagraphup
\paragraph{Similarity among Action Representations.}
In the factored multiplicative interactions, every action is linearly transformed to $f$ factors ($\mathbf{W}^{a}\mathbf{a}$ in Equation~\ref{eq:factorization}). 
In Figure~\ref{fig:action-correlation} we present the cosine similarity between every pair of action-factors 
after training in Seaquest.
`N' and `F' corresponds to `no-operation' and `fire'. 
Arrows correspond to movements with (black) or without (white) `fire'.
There are positive correlations between actions that have the same movement directions (e.g., `up' and `up+fire'), and negative correlations between actions that have opposing directions.  
These results are reasonable and discovered automatically in learning good predictions. 

\textbf{Distinguishing Controlled and Uncontrolled Objects} 
is itself a hard and interesting problem. 
Bellemare et al.~\cite{bellemare2012investigating} proposed a framework to learn \textit{contingent regions} of an image affected by agent action, suggesting that contingency awareness is useful for model-free agents. 
We show that our architectures implicitly learn contingent regions as they learn to predict the entire image.

\begin{wrapfigure}{r}{0.38\textwidth}
\vspace{-0.18in}
 \begin{center}
    \includegraphics[width=0.38\textwidth]{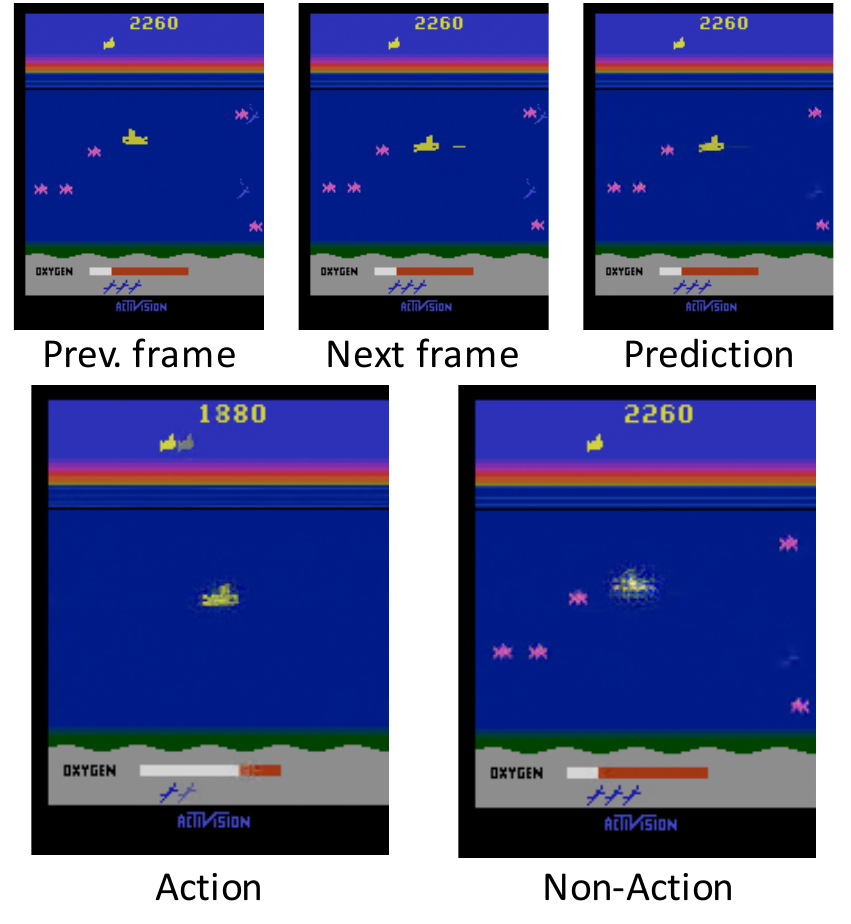}
 \end{center}
 \vspace{-15pt}
 \caption{Distinguishing controlled and uncontrolled objects. \emph{Action} image shows a prediction given only learned action-factors with high variance; 
\emph{Non-Action} image given only low-variance factors.} 
 \vspace{-20pt}
 \label{fig:disentangle}
\end{wrapfigure}
In our architectures, a factor ($f_i=(\mathbf{W}_{i,:}^a)^{\top}\mathbf{a}$) with higher variance 
measured over all possible actions, $\mbox{Var}\left(f_i\right)=\mathbb{E_\textbf{a}}\left[\left(f_{i}-\mathbb{E_\textbf{a}}[{f}_{i}]\right)^2\right]$, is more likely to transform an image differently depending on actions, and so 
we assume such factors are responsible for transforming the parts of the image related to actions.
We therefore collected the high variance (referred to as ``highvar") factors from the model trained on Seaquest (around 40$\%$ of factors), and collected the remaining factors into a low variance (``lowvar") subset.
Given an image and an action, we did two controlled forward propagations: giving only highvar factors (by setting the other factors to zeros) and vice versa.
The results are visualized as `Action' and `Non-Action' in Figure~\ref{fig:disentangle}.
Interestingly, 
given only highvar-factors (Action), the model predicts sharply the movement of the object controlled by actions, while the other parts are mean pixel values. In contrast, given only lowvar-factors (Non-Action), the model predicts the movement of the other objects and the background (e.g., oxygen), and the controlled object stays at its previous location. 
This result implies that our model learns to distinguish between controlled objects and uncontrolled objects and transform them using disentangled representations (see \cite{rifai2012disentangling,icml2014disentangling,yang2015weaksupdisentangle} for related work on disentangling factors of variation).

\cutsectionup 
\section{Conclusion} 
\cutsectiondown
\vspace{-0.01in}
This paper introduced two different novel deep architectures that predict future frames that are dependent on actions and showed qualitatively and quantitatively that they are able to predict visually-realistic and useful-for-control frames over 100-step futures on several Atari game domains. To our knowledge, this is the first paper to show good deep predictions in Atari games. 
Since our architectures were domain independent we expect that they will generalize to many vision-based RL problems. 
In future work we will learn models that predict future reward in addition to predicting future frames and evaluate the performance of our architectures in model-based RL. 

\cutparagraphup
\paragraph{Acknowledgments.}
This work was supported by NSF grant IIS-1526059, Bosch Research, and ONR grant N00014-13-1-0762. Any opinions, findings, conclusions, or recommendations expressed here are those of the authors and do not necessarily reflect the views of the sponsors.

\vspace{-0.04in}
\cutsectionup
\bibliographystyle{abbrv}
\footnotesize{
\bibliography{references}

\begin{thebibliography}{10}

\bibitem{bellemare13arcade}
M.~G. {Bellemare}, Y.~{Naddaf}, J.~{Veness}, and M.~{Bowling}.
\newblock The {Arcade Learning Environment}: An evaluation platform for general
  agents.
\newblock {\em Journal of Artificial Intelligence Research}, 47:253--279, 2013.

\bibitem{bellemare2012investigating}
M.~G. Bellemare, J.~Veness, and M.~Bowling.
\newblock Investigating contingency awareness using {Atari} 2600 games.
\newblock In {\em AAAI}, 2012.

\bibitem{bellemare2013bayesian}
M.~G. Bellemare, J.~Veness, and M.~Bowling.
\newblock Bayesian learning of recursively factored environments.
\newblock In {\em ICML}, 2013.

\bibitem{bellemare2014skip}
M.~G. Bellemare, J.~Veness, and E.~Talvitie.
\newblock Skip context tree switching.
\newblock In {\em ICML}, 2014.

\bibitem{bengio2009learning}
Y.~Bengio.
\newblock Learning deep architectures for {AI}.
\newblock {\em Foundations and Trends in Machine Learning}, 2(1):1--127, 2009.

\bibitem{bengio2009curriculum}
Y.~Bengio, J.~Louradour, R.~Collobert, and J.~Weston.
\newblock Curriculum learning.
\newblock In {\em ICML}, 2009.

\bibitem{ciresan2012multi}
D.~Ciresan, U.~Meier, and J.~Schmidhuber.
\newblock Multi-column deep neural networks for image classification.
\newblock In {\em CVPR}, 2012.

\bibitem{dosovitskiy2014learning}
A.~Dosovitskiy, J.~T. Springenberg, and T.~Brox.
\newblock Learning to generate chairs with convolutional neural networks.
\newblock In {\em CVPR}, 2015.

\bibitem{girshick2014rich}
R.~Girshick, J.~Donahue, T.~Darrell, and J.~Malik.
\newblock Rich feature hierarchies for accurate object detection and semantic
  segmentation.
\newblock In {\em CVPR}, 2014.

\bibitem{graves2013generating}
A.~Graves.
\newblock Generating sequences with recurrent neural networks.
\newblock {\em arXiv preprint arXiv:1308.0850}, 2013.

\bibitem{guo2014deep}
X.~Guo, S.~Singh, H.~Lee, R.~L. Lewis, and X.~Wang.
\newblock Deep learning for real-time {Atari} game play using offline
  {Monte-Carlo} tree search planning.
\newblock In {\em NIPS}, 2014.

\bibitem{hochreiter1997long}
S.~Hochreiter and J.~Schmidhuber.
\newblock Long short-term memory.
\newblock {\em {Neural Computation}}, 9(8):1735--1780, 1997.

\bibitem{jia2014caffe}
Y.~Jia, E.~Shelhamer, J.~Donahue, S.~Karayev, J.~Long, R.~Girshick,
  S.~Guadarrama, and T.~Darrell.
\newblock Caffe: Convolutional architecture for fast feature embedding.
\newblock In {\em ACM Multimedia}, 2014.

\bibitem{karpathy2014large}
A.~Karpathy, G.~Toderici, S.~Shetty, T.~Leung, R.~Sukthankar, and L.~Fei-Fei.
\newblock Large-scale video classification with convolutional neural networks.
\newblock In {\em CVPR}, 2014.

\bibitem{kocsis2006bandit}
L.~Kocsis and C.~Szepesv{\'a}ri.
\newblock Bandit based {Monte-Carlo} planning.
\newblock In {\em ECML}. 2006.

\bibitem{krizhevsky2012imagenet}
A.~Krizhevsky, I.~Sutskever, and G.~E. Hinton.
\newblock Imagenet classification with deep convolutional neural networks.
\newblock In {\em NIPS}, 2012.

\bibitem{lenz_deepMPC_2015}
I.~Lenz, R.~Knepper, and A.~Saxena.
\newblock Deep{MPC}: Learning deep latent features for model predictive
  control.
\newblock In {\em RSS}, 2015.

\bibitem{memisevic2013learning}
R.~Memisevic.
\newblock Learning to relate images.
\newblock {\em IEEE TPAMI}, 35(8):1829--1846, 2013.

\bibitem{michalski2014modeling}
V.~Michalski, R.~Memisevic, and K.~Konda.
\newblock Modeling deep temporal dependencies with recurrent grammar cells.
\newblock In {\em NIPS}, 2014.

\bibitem{mittelman2014structured}
R.~Mittelman, B.~Kuipers, S.~Savarese, and H.~Lee.
\newblock Structured recurrent temporal restricted {Boltzmann} machines.
\newblock In {\em ICML}, 2014.

\bibitem{mnih2013playing}
V.~Mnih, K.~Kavukcuoglu, D.~Silver, A.~Graves, I.~Antonoglou, D.~Wierstra, and
  M.~Riedmiller.
\newblock Playing {Atari} with deep reinforcement learning.
\newblock {\em arXiv preprint arXiv:1312.5602}, 2013.

\bibitem{mnih2015human}
V.~Mnih, K.~Kavukcuoglu, D.~Silver, A.~A. Rusu, J.~Veness, M.~G. Bellemare,
  A.~Graves, M.~Riedmiller, A.~K. Fidjeland, G.~Ostrovski, et~al.
\newblock Human-level control through deep reinforcement learning.
\newblock {\em Nature}, 518(7540):529--533, 2015.

\bibitem{nair2010rectified}
V.~Nair and G.~E. Hinton.
\newblock Rectified linear units improve restricted {Boltzmann} machines.
\newblock In {\em ICML}, 2010.

\bibitem{icml2014disentangling}
S.~Reed, K.~Sohn, Y.~Zhang, and H.~Lee.
\newblock Learning to disentangle factors of variation with manifold
  interaction.
\newblock In {\em ICML}, 2014.

\bibitem{rifai2012disentangling}
S.~Rifai, Y.~Bengio, A.~Courville, P.~Vincent, and M.~Mirza.
\newblock Disentangling factors of variation for facial expression recognition.
\newblock In {\em ECCV}. 2012.

\bibitem{schmidhuber2015deep}
J.~Schmidhuber.
\newblock Deep learning in neural networks: An overview.
\newblock {\em Neural Networks}, 61:85--117, 2015.

\bibitem{schmidhuber1991learning}
J.~Schmidhuber and R.~Huber.
\newblock Learning to generate artificial fovea trajectories for target
  detection.
\newblock {\em International Journal of Neural Systems}, 2:125--134, 1991.

\bibitem{srivastava2015unsupervised}
N.~Srivastava, E.~Mansimov, and R.~Salakhutdinov.
\newblock Unsupervised learning of video representations using {LSTM}s.
\newblock In {\em ICML}, 2015.

\bibitem{sutskever2009recurrent}
I.~Sutskever, G.~E. Hinton, and G.~W. Taylor.
\newblock The recurrent temporal restricted {Boltzmann} machine.
\newblock In {\em NIPS}, 2009.

\bibitem{sutskever2011generating}
I.~Sutskever, J.~Martens, and G.~E. Hinton.
\newblock Generating text with recurrent neural networks.
\newblock In {\em ICML}, 2011.

\bibitem{sutskever2014sequence}
I.~Sutskever, O.~Vinyals, and Q.~Le.
\newblock Sequence to sequence learning with neural networks.
\newblock In {\em NIPS}, 2014.

\bibitem{szegedy2014going}
C.~Szegedy, W.~Liu, Y.~Jia, P.~Sermanet, S.~Reed, D.~Anguelov, D.~Erhan,
  V.~Vanhoucke, and A.~Rabinovich.
\newblock Going deeper with convolutions.
\newblock {\em arXiv preprint arXiv:1409.4842}, 2014.

\bibitem{taylor2009factored}
G.~W. Taylor and G.~E. Hinton.
\newblock Factored conditional restricted {Boltzmann} machines for modeling
  motion style.
\newblock In {\em ICML}, 2009.

\bibitem{tieleman2012}
T.~Tieleman and G.~Hinton.
\newblock Lecture 6.5 - {RMSProp}: Divde the gradient by a running average of
  its recent magnitude.
\newblock {\em Coursera}, 2012.

\bibitem{DBLP:journals/corr/TranBFTP14}
D.~Tran, L.~Bourdev, R.~Fergus, L.~Torresani, and M.~Paluri.
\newblock Learning spatiotemporal features with {3D} convolutional networks.
\newblock In {\em ICCV}, 2015.

\bibitem{watkins1992q}
C.~J. Watkins and P.~Dayan.
\newblock Q-learning.
\newblock {\em Machine learning}, 8(3-4):279--292, 1992.

\bibitem{yang2015weaksupdisentangle}
J.~Yang, S.~Reed, M.-H. Yang, and H.~Lee.
\newblock Weakly-supervised disentangling with recurrent transformations for
  3{D} view synthesis.
\newblock In {\em NIPS}, 2015.

\end{thebibliography}
}
\clearpage
\appendix
\section{Network Architectures and Training Details}
\begin{figure}[H]
  \centering
  \begin{subfigure}{1.0\textwidth}
	  \centering
	  \includegraphics[height=0.08\textheight]{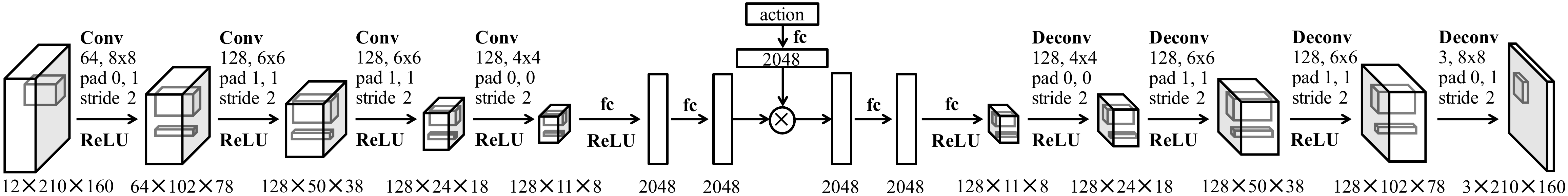}
	  \caption{Feedforward Architecture}
	  \vspace{15pt}
  \end{subfigure}
  \begin{subfigure}{1.0\textwidth}
	  \centering
	  \includegraphics[height=0.08\textheight]{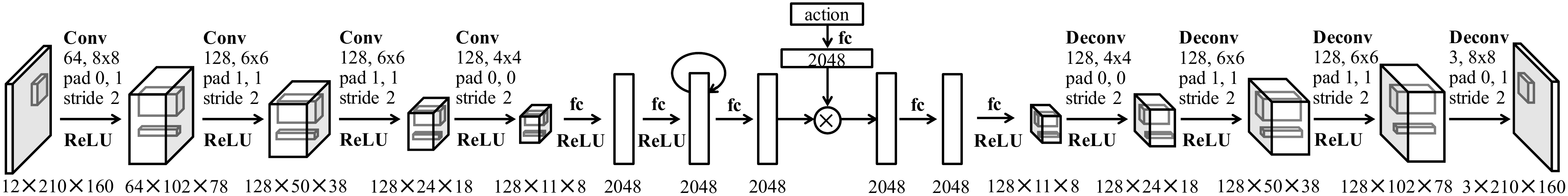}
	  \caption{Recurrent Architecture}
	  \vspace{15pt}
  \end{subfigure}
  \begin{subfigure}{1.0\textwidth}
	  \centering
	  \includegraphics[height=0.07\textheight]{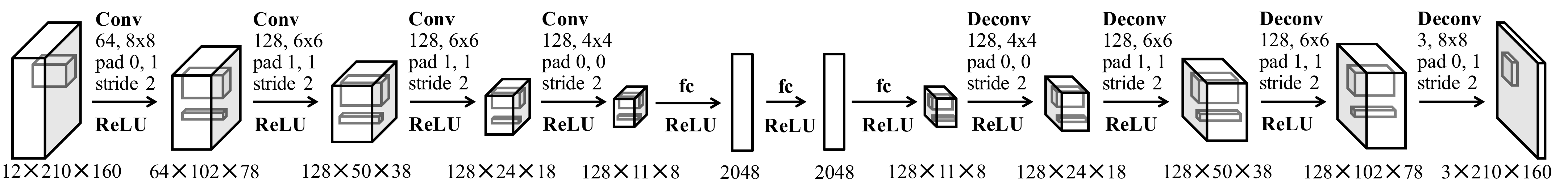}
	  \caption{No-action Feedforward}
	  \vspace{15pt}
  \end{subfigure}
  \begin{subfigure}{1.0\textwidth}
	  \centering
	  \includegraphics[height=0.08\textheight]{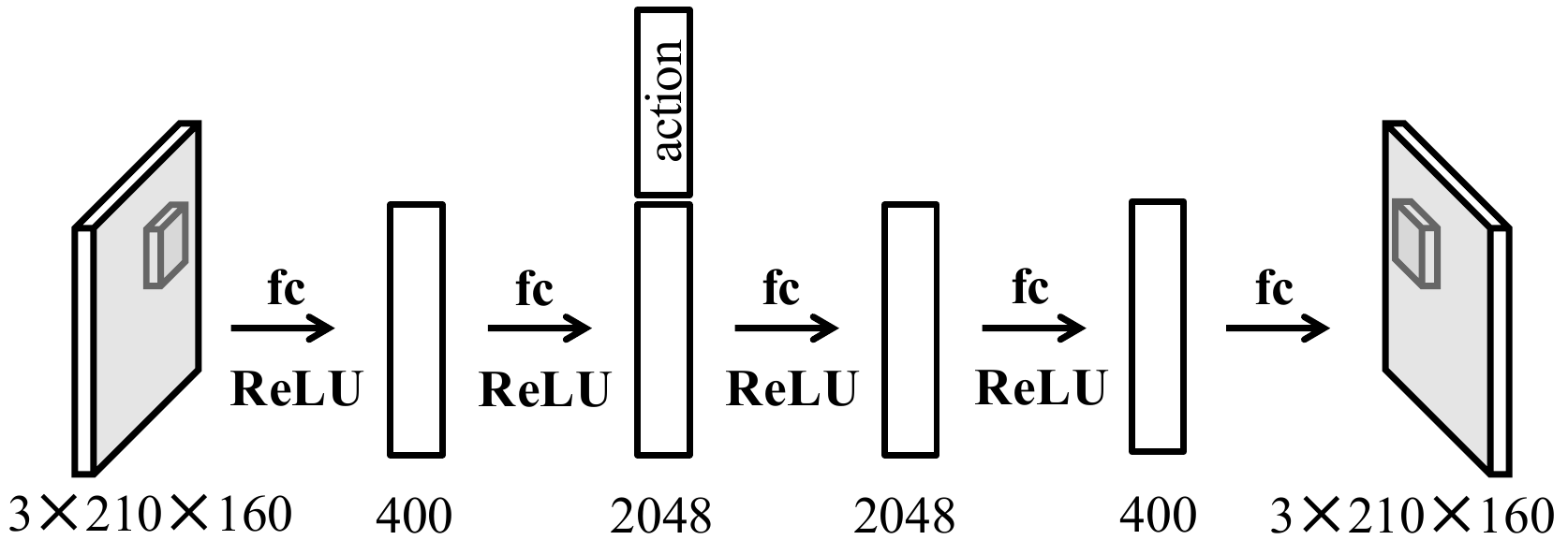}
	  \caption{MLP}
  \end{subfigure}
\caption{Network architectures. `$\times$' indicates element-wise multiplication. The text in each (de-)convolution layer describes the number of filters, the size of the kernel, padding (height and width), and stride.}
\label{fig:architecture-rgb}
\end{figure}

The network architectures of the proposed models and the baselines are illustrated in Figure~\ref{fig:architecture-rgb}. 

The weight of LSTM is initialized from a uniform distribution of $[-0.08,0.08]$. The weight of the fully-connected layer from the encoded feature to the factored layer and from the action to the factored layer are initialized from a uniform distribution of $[-1,1]$ and $[-0.1,0.1]$ respectively.

The total number of iterations is $1.5\times10^6$, $10^6$, and $10^6$ for each training phase (1-step, 3-step, and 5-step). 
The learning rate is multiplied by $0.9$ after every $10^5$ iterations. 

\clearpage
\section{Informed Exploration}
The entire DQN algorithm with informed exploration is described in Algorithm~\ref{alg-exploration}.
\begin{algorithm}
\small
\caption{Deep Q-learning with informed exploration \label{alg-exploration}}
\begin{algorithmic}
\State Allocate capacity of replay memory $R$
\State \textbf{Allocate capacity of trajectory memory $D$}
\State Initialize parameters $\theta$ of DQN 
\While{$steps<M$}
\State Reset game and observe image $x_{1}$
\State \textbf{Store image $x_{1}$ in $D$}
\For{$t$=1 to $T$}
\State Sample $c$ from Bernoulli distribution with parameter $\epsilon$
\State Set $a_t=\begin{cases} \operatorname{\textbf{argmin}}_{a}n_{D}\left(x_t^{(a)}\right) &\mbox{if } c=1 \\
       \operatorname{argmax}_{a}Q\left(\phi\left(s_t\right),a;\theta\right)) & \mbox{otherwise } \end{cases}$
\State Choose action $a_t$, observe reward $r_t$ and image $x_{t+1}$
\State Set $s_{t+1}=x_{t-2:t+1}$ and preprocess images $\phi_{t+1}=\phi\left(s_{t+1}\right)$
\State \textbf{Store image $x_{t+1}$ in $D$}
\State Store transition $\left(\phi_t,a_t,r_t,\phi_{t+1}\right)$ in $R$
\State Sample a mini-batch of transitions $\left\{\phi_j,a_j,r_j,\phi_{j+1}\right\}$ from $R$
\State Update $\theta$ based on the mini-batch and Bellman equation
\State $steps = steps + 1$
\EndFor
\EndWhile
\end{algorithmic}
\end{algorithm}
\vspace{-15pt}
\begin{figure}[H]
  \centering
  \includegraphics[width=\linewidth]{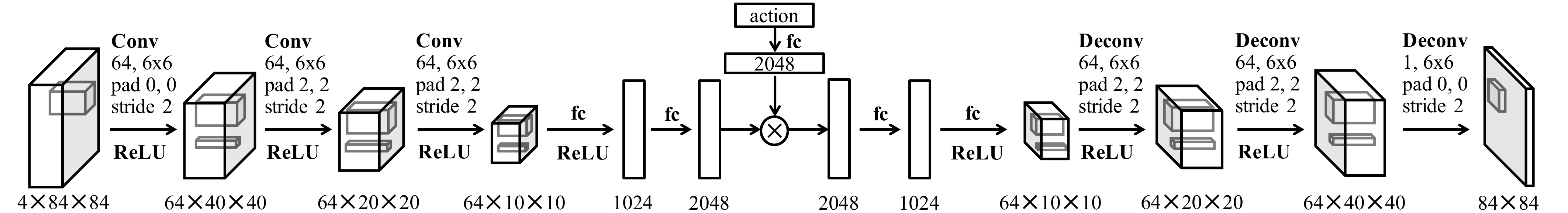}
  \caption{Feedforward encoding network for gray-scaled and down-sampled images. }
  \label{fig:architecture-gray}
  \vspace{-10pt}
\end{figure}

\cutparagraphup
\paragraph{Predictive Model for Informed Exploration.}
A feedforward encoding network (illustrated in Figure~\ref{fig:architecture-gray}) trained on down-sampled and gray-scaled images is used for computational efficiency.
We trained the model on 1-step prediction objective with learning rate of $10^{-4}$ and batch size of $32$. The pixel values are subtracted by mean pixel values and divided by 128.
RMSProp is used with momentum of $0.9$, (squared) gradient momentum of $0.95$, and min squared gradient of $0.01$. 

\begin{figure}[H]
  \centering
  \begin{subfigure}{0.23\textwidth}
	  \centering
	  \includegraphics[width=\linewidth]{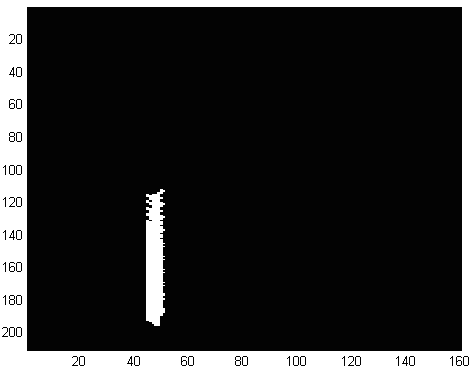}
	  \caption{Random}
  \end{subfigure}
  \begin{subfigure}{0.23\textwidth}
	  \centering
	  \includegraphics[width=\linewidth]{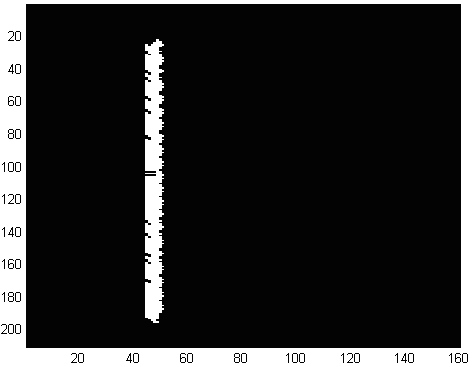}
	  \caption{Informed}
  \end{subfigure}
  \begin{subfigure}{0.23\textwidth}
	  \centering
	  \includegraphics[width=\linewidth]{exp-pac_rand.png}
	  \caption{Random}
  \end{subfigure}
  \begin{subfigure}{0.23\textwidth}
	  \centering	
	  \includegraphics[width=\linewidth]{exp-pac_exp.png}
	  \caption{Informed}
  \end{subfigure}
\caption{Comparison between two exploration methods on Freeway (Left) and Ms Pacman (Right). Each heat map shows the trajectories of the agent measured from 2500 steps from each exploration strategy. }
\label{fig:heatmap-all}
\end{figure}

\cutparagraphup
\paragraph{Comparison to Random Exploration.}
Figure~\ref{fig:heatmap-all} visualizes the difference between random exploration and informed exploration in two games. In Freeway, where the agent gets rewards by reaching the top lane, the agent moves only around the bottom area in the random exploration, which results in $4.6\times10^5$ steps to get the first reward. 
On the other hand, the agent moves around all locations in the informed exploration and receives the first reward in $86$ steps. The similar result is found in Ms Pacman. 

\begin{figure}[H]
  \centering
  \begin{subfigure}{0.27\textwidth}
	  \centering
	  \includegraphics[width=\linewidth]{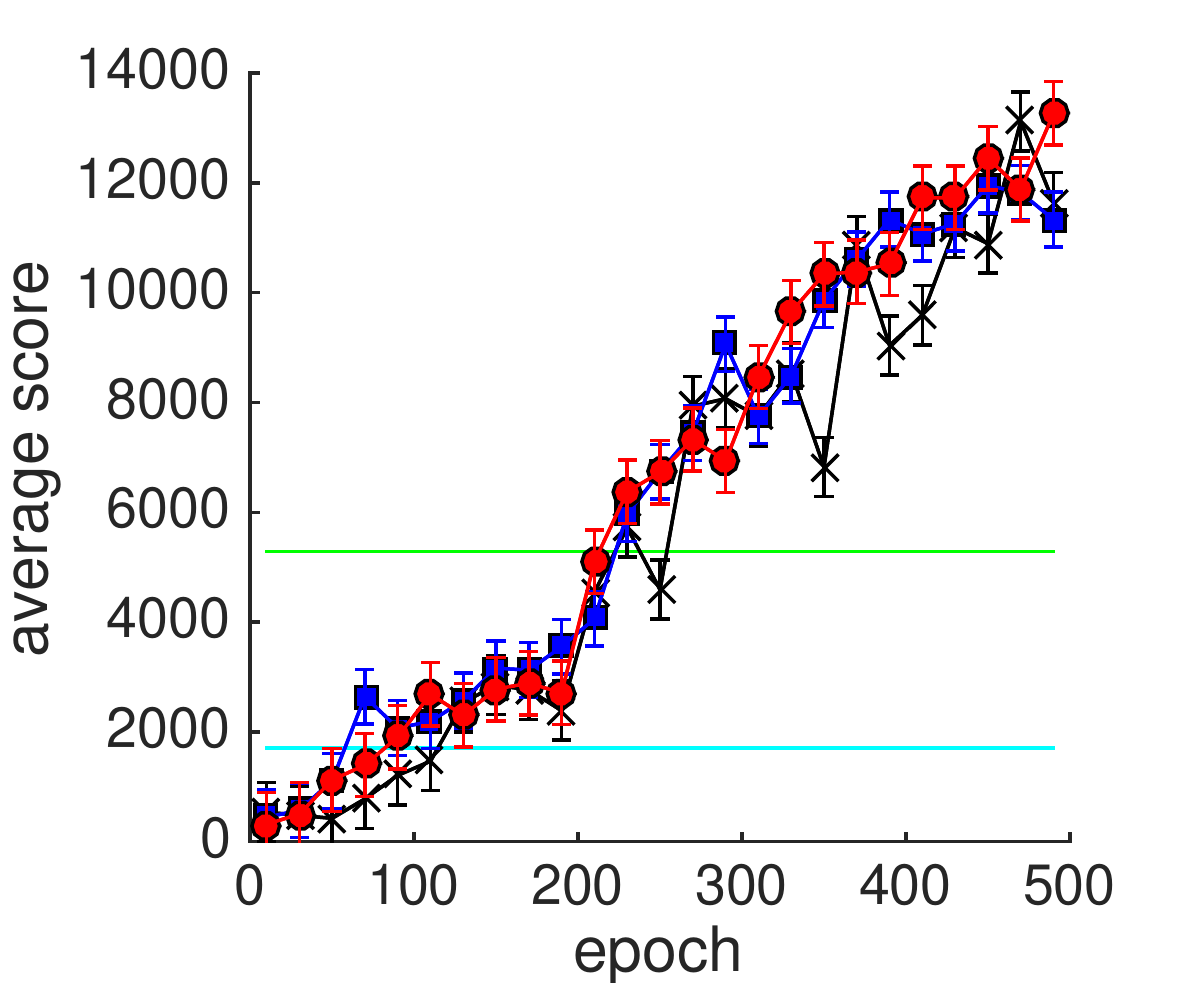}
	  \caption{Seaquest}
  \end{subfigure}
  \begin{subfigure}{0.27\textwidth}
	  \centering
	  \includegraphics[width=\linewidth]{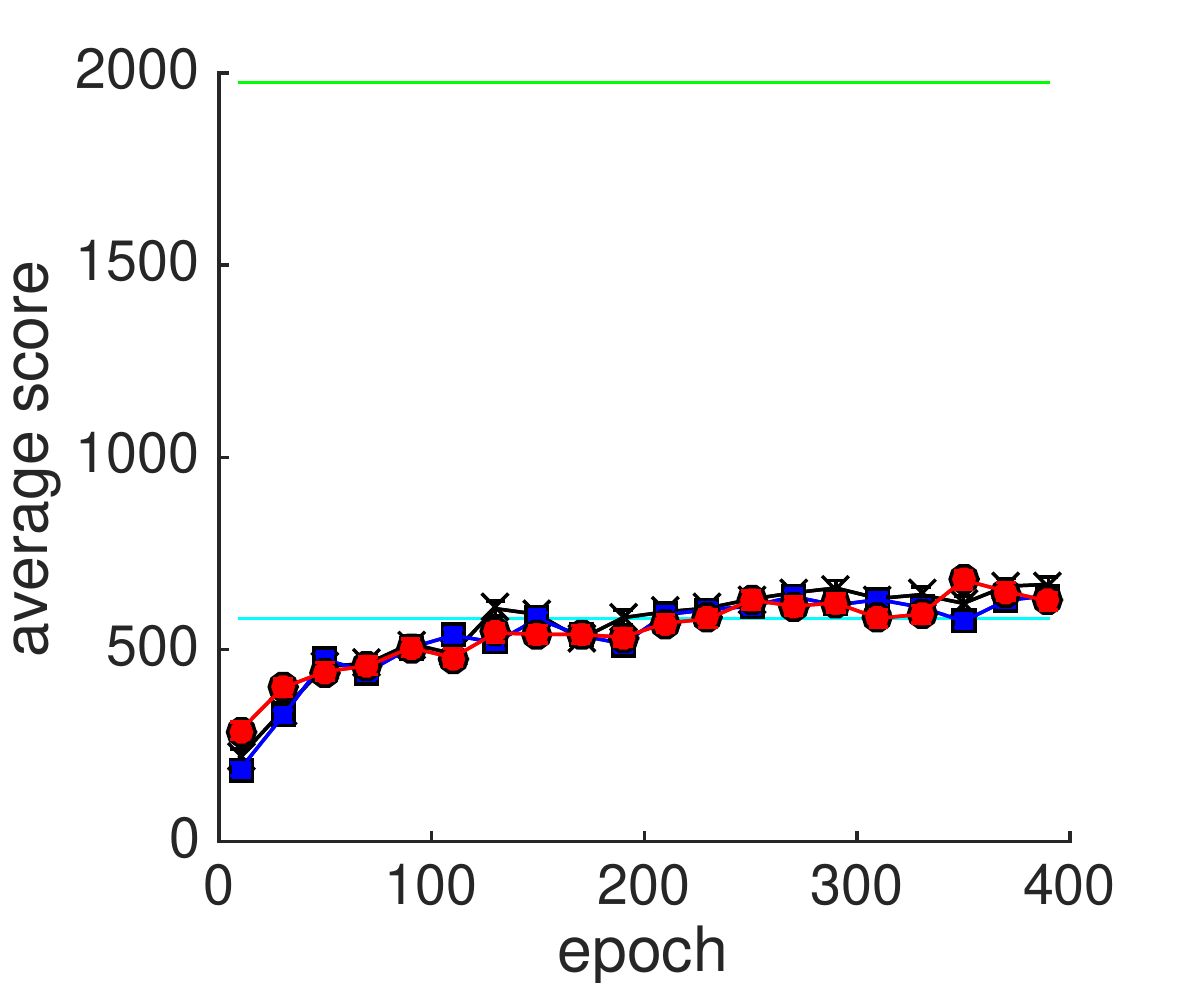}
    \caption{Space Invaders}
  \end{subfigure}
  \begin{subfigure}{0.27\textwidth}
	  \centering
	  \includegraphics[width=\linewidth]{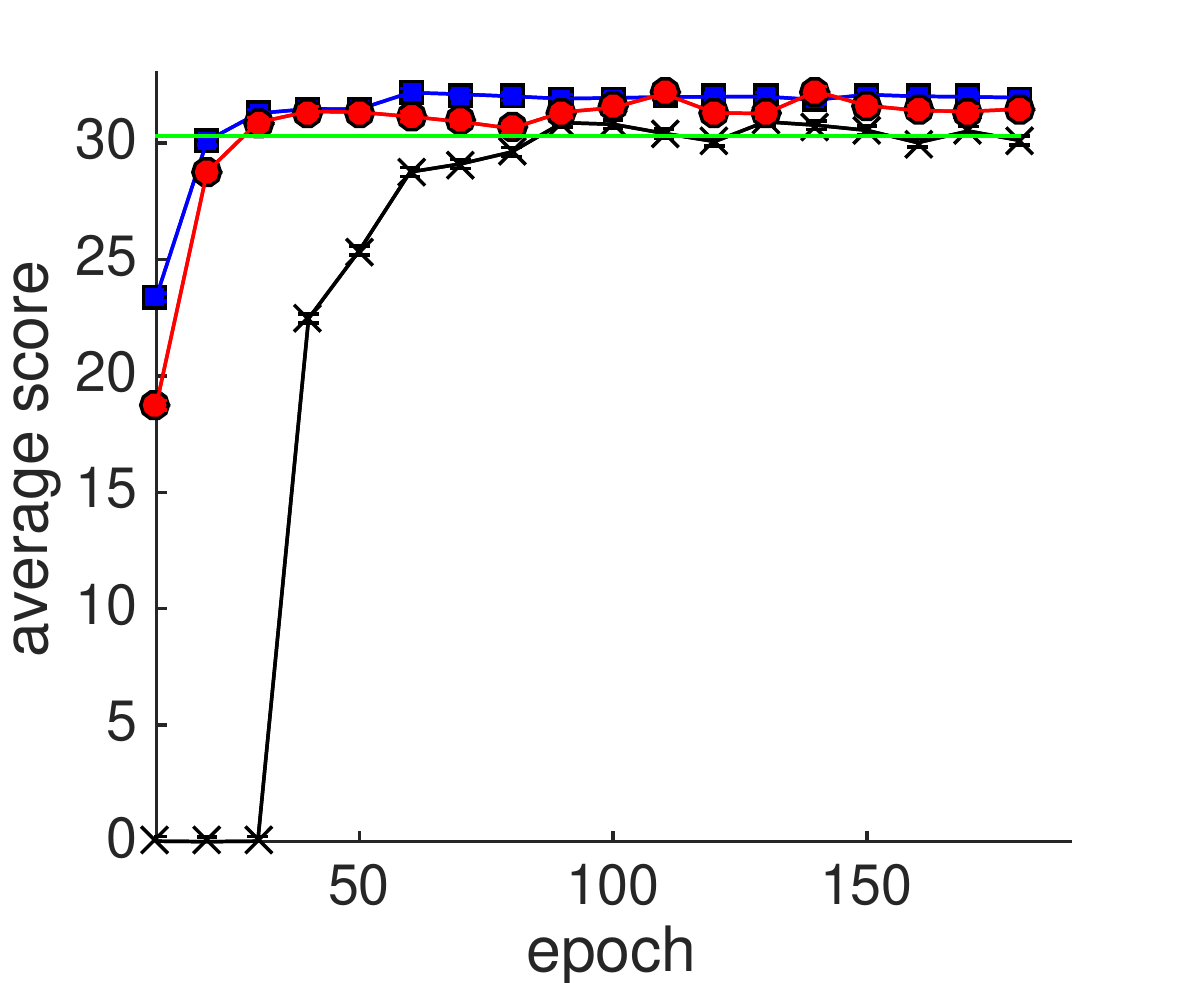}
      \caption{Freeway}
  \end{subfigure}
  \begin{subfigure}{0.27\textwidth}
	  \centering	
	  \includegraphics[width=\linewidth]{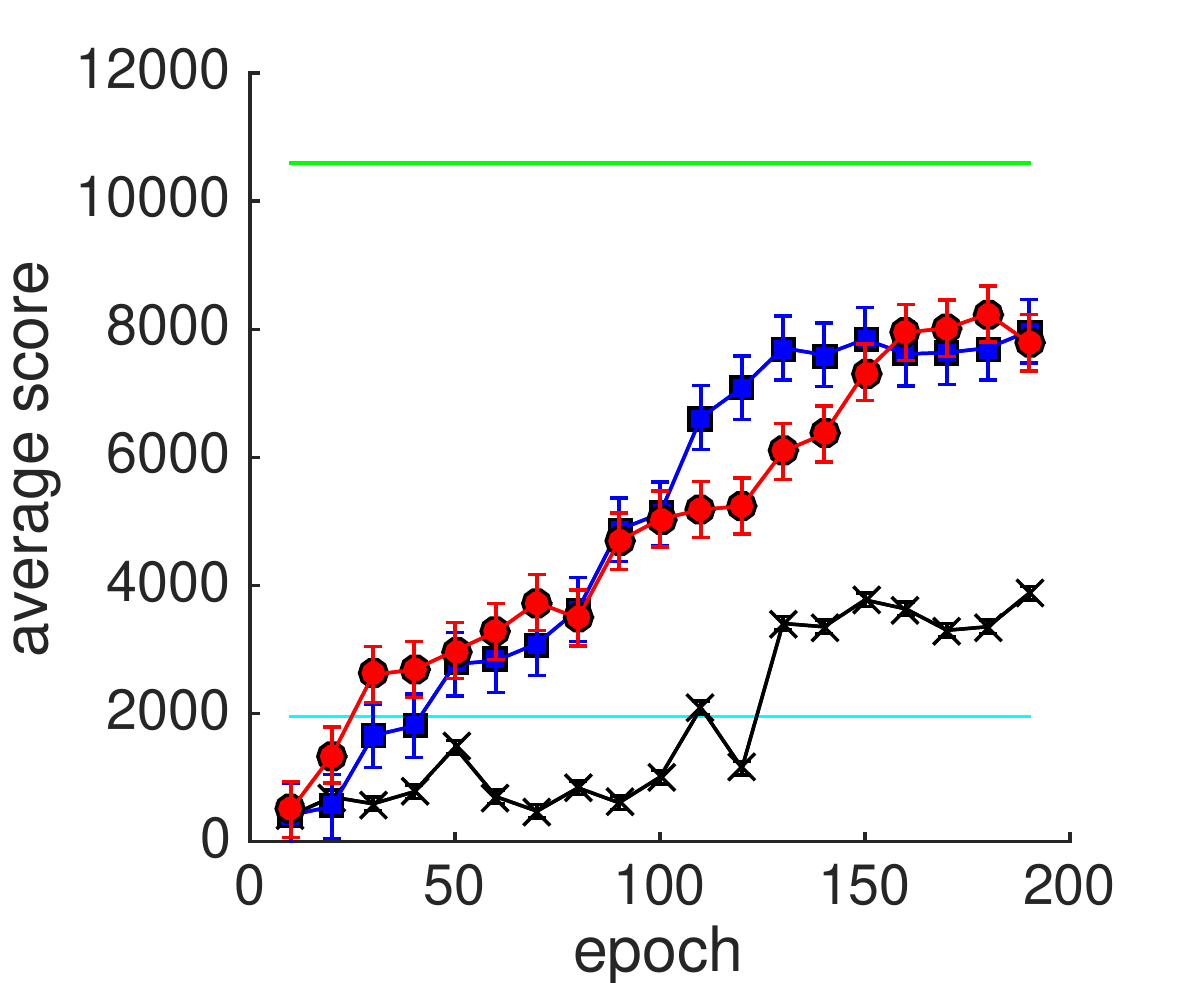}
      \caption{QBert}
  \end{subfigure}
  \begin{subfigure}{0.27\textwidth}
	  \centering
	  \includegraphics[width=\linewidth]{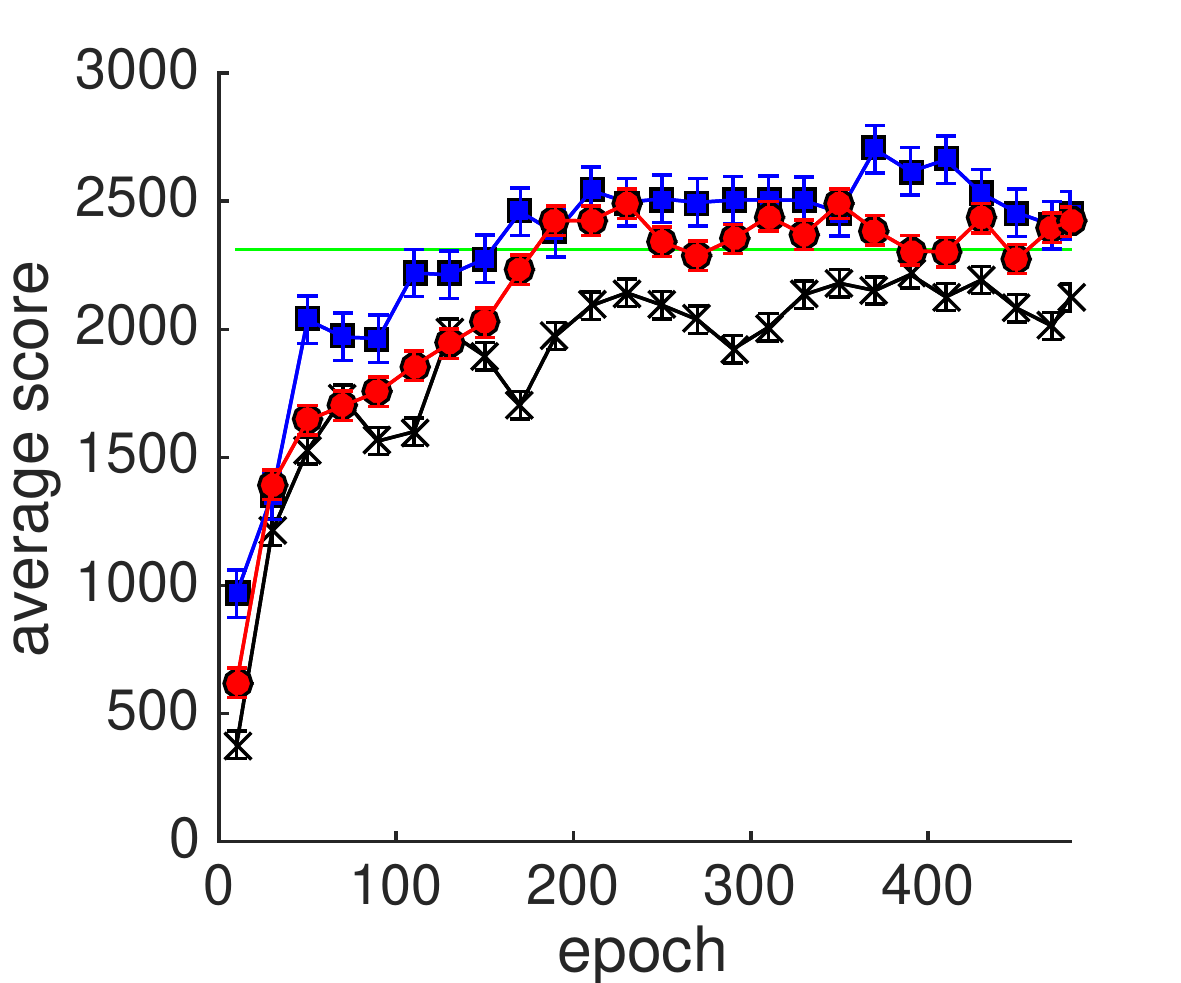}
      \caption{Ms Pacman}
  \end{subfigure}
    \begin{subfigure}{0.27\textwidth}
	  \centering
	  \includegraphics[width=\linewidth]{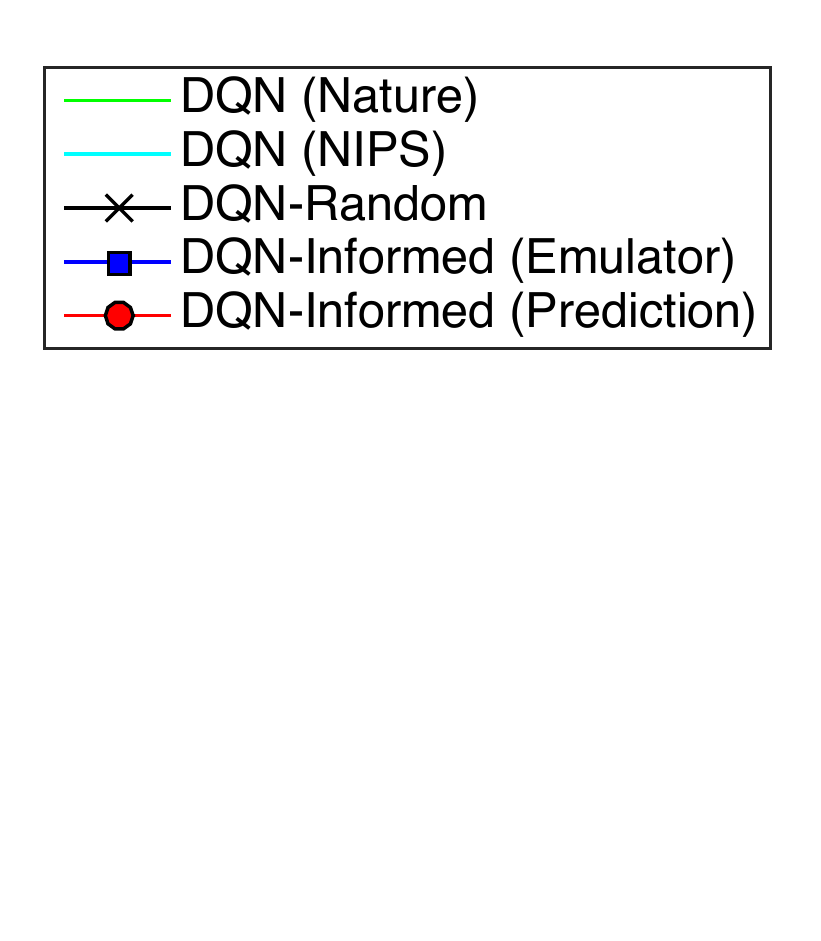}
  \end{subfigure}
\caption{Learning curves of DQNs with standard errors. The red and blue curves are informed exploration using our predictive model and the emulator respectively. The black curves are DQNs with random exploration. The average game score is measured from 100 game plays with $\epsilon$-greedy policy with $\epsilon=0.05$. }
\label{fig:informedexp}
\vspace{-10pt}
\end{figure}

\cutparagraphup
\paragraph{Application to Deep Q-learning.}
The results of the informed exploration using the game emulator and our predictive model are reported in Figure~\ref{fig:informedexp} and Table~\ref{table:game-score}.
Our DQN replication follows~\cite{mnih2013playing}, which uses a smaller CNN than~\cite{mnih2015human}. 

\begin{table}[H]
\centering
\small
\setlength{\tabcolsep}{4pt}
\begin{tabular}{lccccc}
\toprule
Model & Seaquest & S. Invaders & Freeway & QBert & Ms Pacman \\ 
\midrule
DQN (Nature)~\cite{mnih2015human} 	& 5286 & 1976 & 30.3 & 10596 & 2311 \\ 
DQN (NIPS)~\cite{mnih2013playing} 		& 1705 & 581 & - & 1952 & - \\ 
\midrule
Our replication of ~\cite{mnih2013playing} & 13119 (538) & 698 (20) & 30.9 (0.2) & 3876 (106) & 2281 (53) \\ 
I.E (Prediction) & 13265 (577) & 681 (23) & 32.2 (0.2) & 8238 (498) & 2522 (57) \\ 
I.E (Emulator)   & 13002 (498) & 708 (17) & 32.2 (0.2) & 7969 (496) & 2702 (92) \\ 
\bottomrule
\end{tabular}
\caption{\small{Average game score with standard error. `I.E' indicates DQN combined with the informed exploration method. `Emulator' and `Prediction' correspond to the emulator and our predictive model for computing $\textbf{x}^{(a)}_{t}$. }} 
\label{table:game-score}
\vspace{-10pt}
\end{table}

\section{Correlation between Actions}
\begin{figure}[H]
    \centering
    \includegraphics[width=0.95\textwidth]{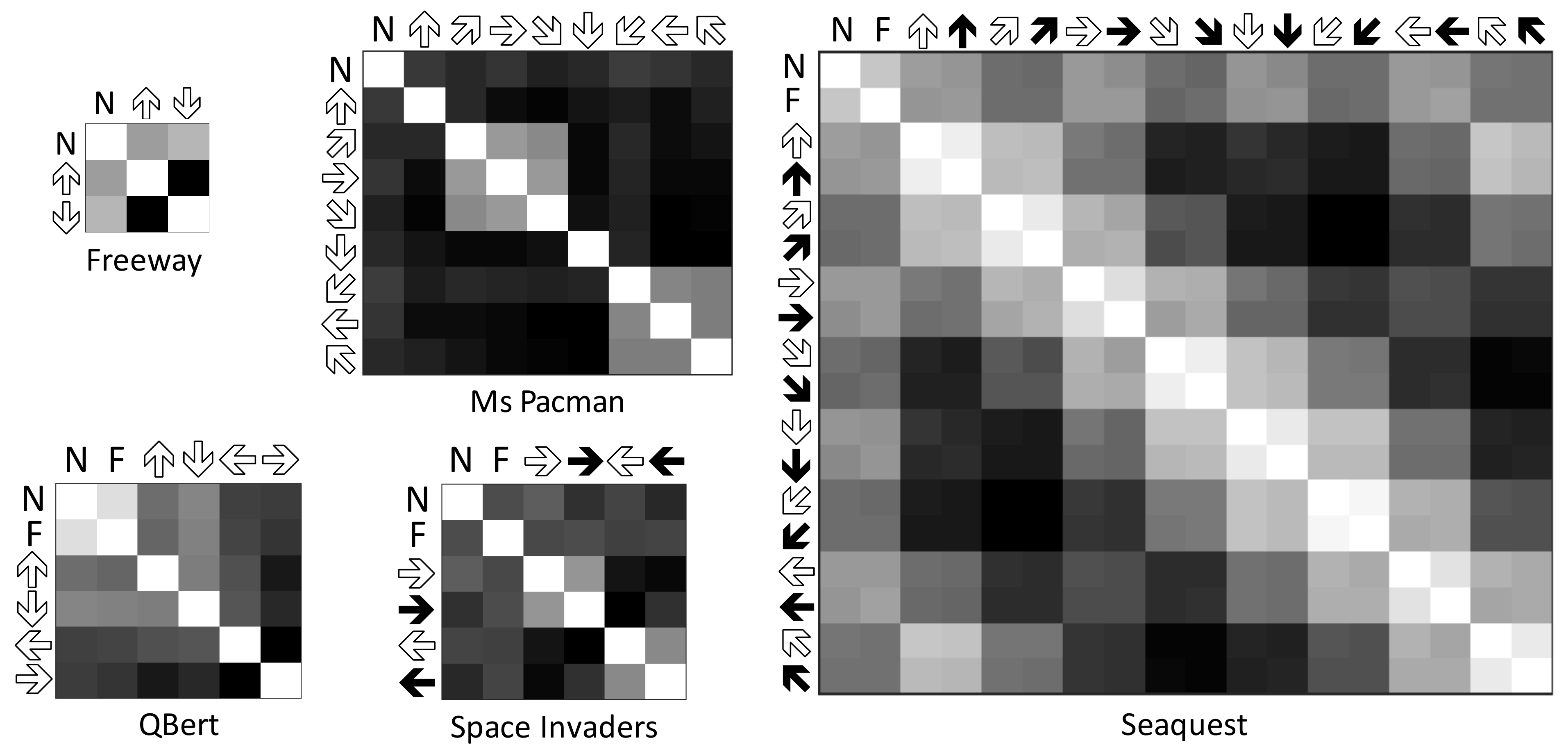}
	\caption{Correlations between actions. The brightness represents consine similarity between pairs of factors.}
	\vspace{-10pt}
\end{figure}

\newpage
\section{Handling Different Actions}
\begin{figure}[ht]
	\begin{subfigure}{\textwidth}
		\centering
		\includegraphics[width=0.95\linewidth]{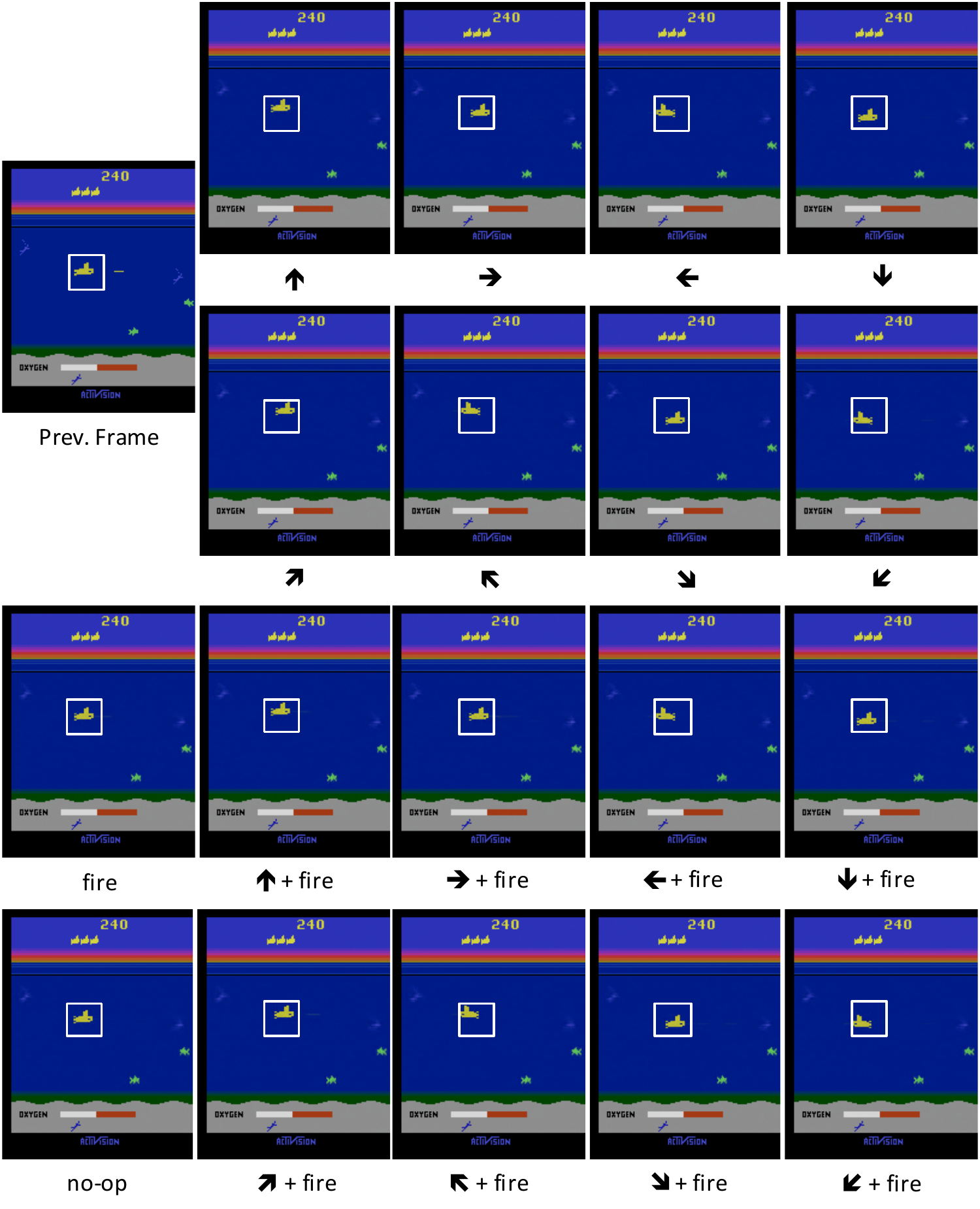} 
		\caption{Seaquest}
	\end{subfigure}
	\vfill
\end{figure}
\begin{figure}[t]
	\centering
	\ContinuedFloat
	\begin{subfigure}{\textwidth}
		\centering
		\includegraphics[width=0.85\linewidth]{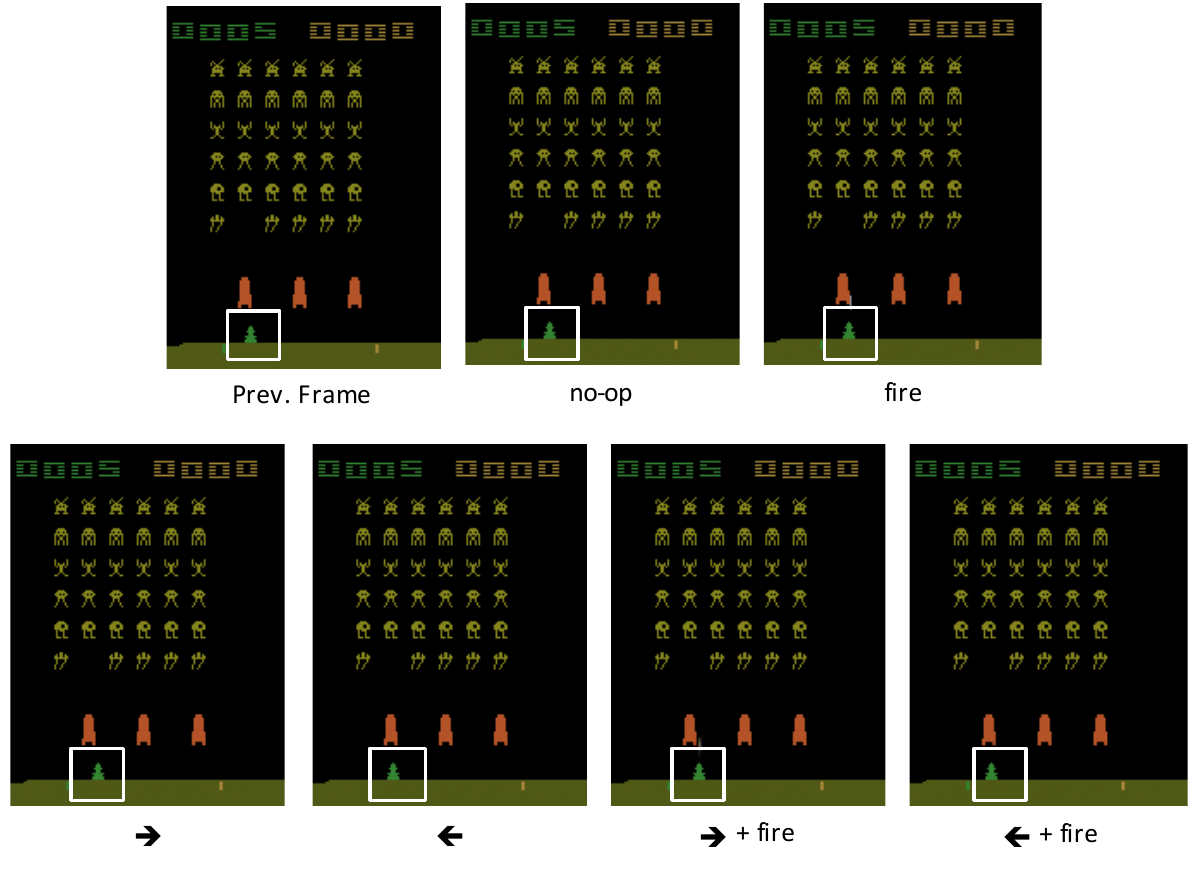} 
		\vspace{-10pt}
		\caption{Space Invaders}
		\vspace{5pt}
	\end{subfigure}
	\begin{subfigure}{\textwidth}
		\centering
		\includegraphics[width=0.85\linewidth]{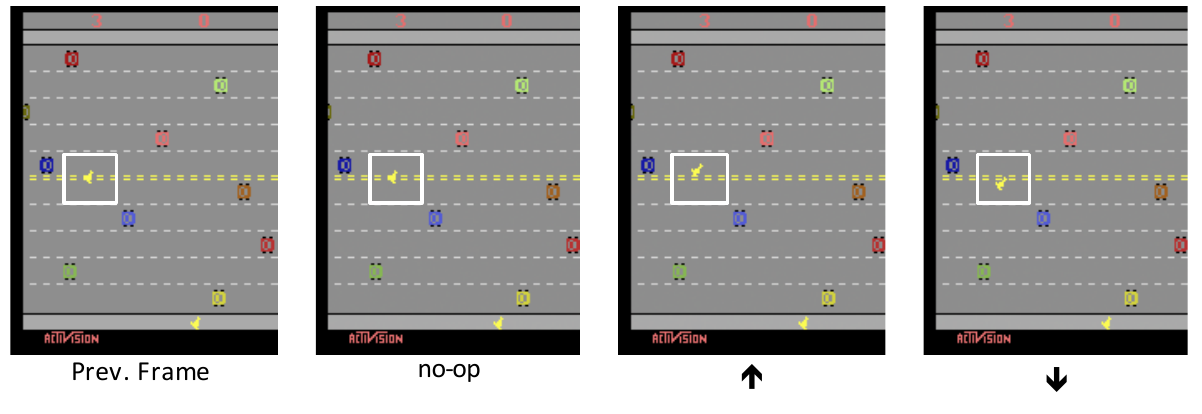} 
		\vspace{-10pt}		
		\caption{Freeway}
		\vspace{5pt}
	\end{subfigure}
	\begin{subfigure}{\textwidth}
		\centering
		\includegraphics[width=0.85\linewidth]{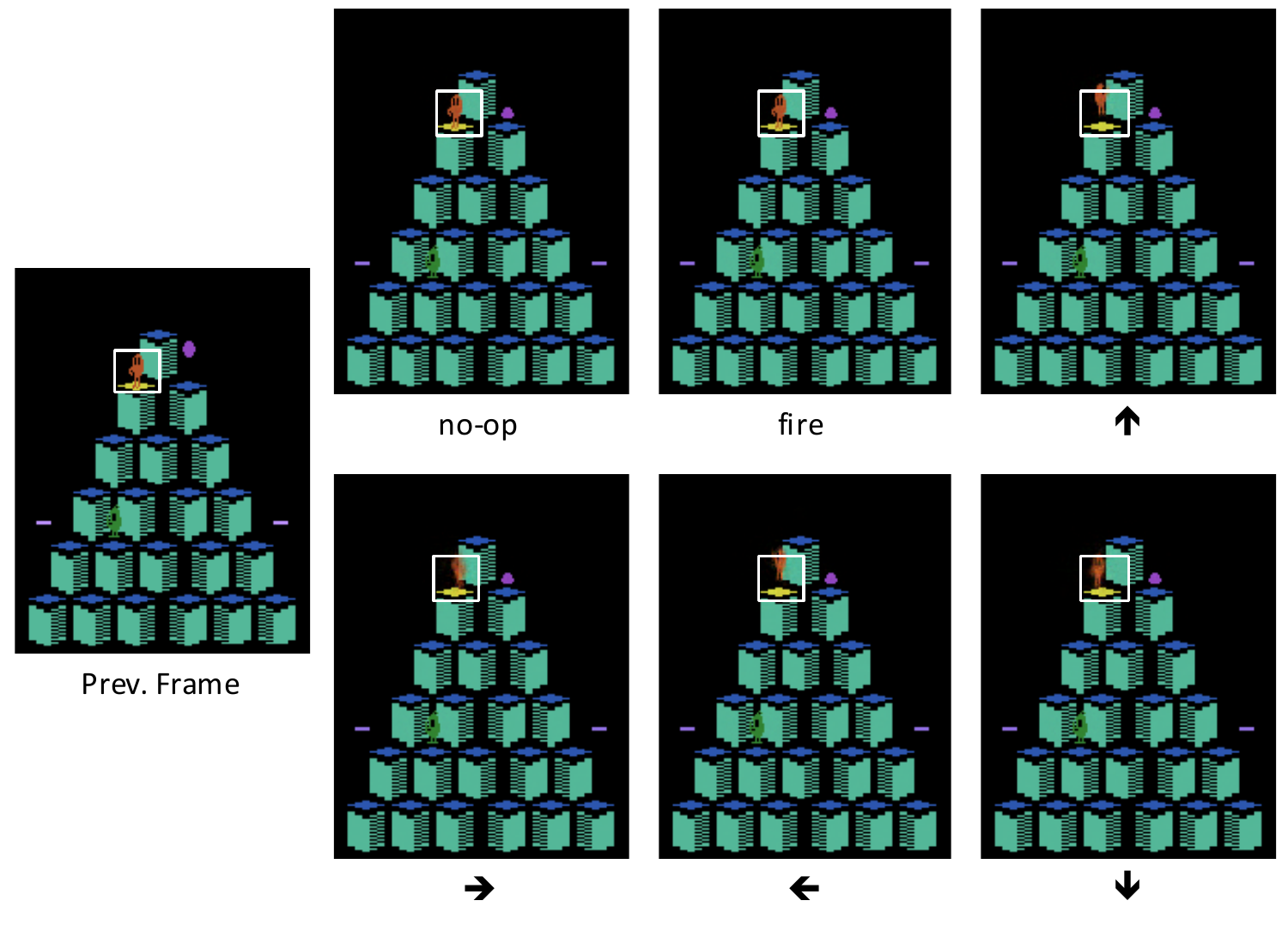} 
		\vspace{-10pt}
		\caption{QBert}
		\vspace{5pt}
	\end{subfigure}
	\vspace{-10pt}
\end{figure}
\clearpage
\begin{figure}[t]
	\centering
	\ContinuedFloat
	\begin{subfigure}{\textwidth}
		\centering
		\includegraphics[width=0.85\linewidth]{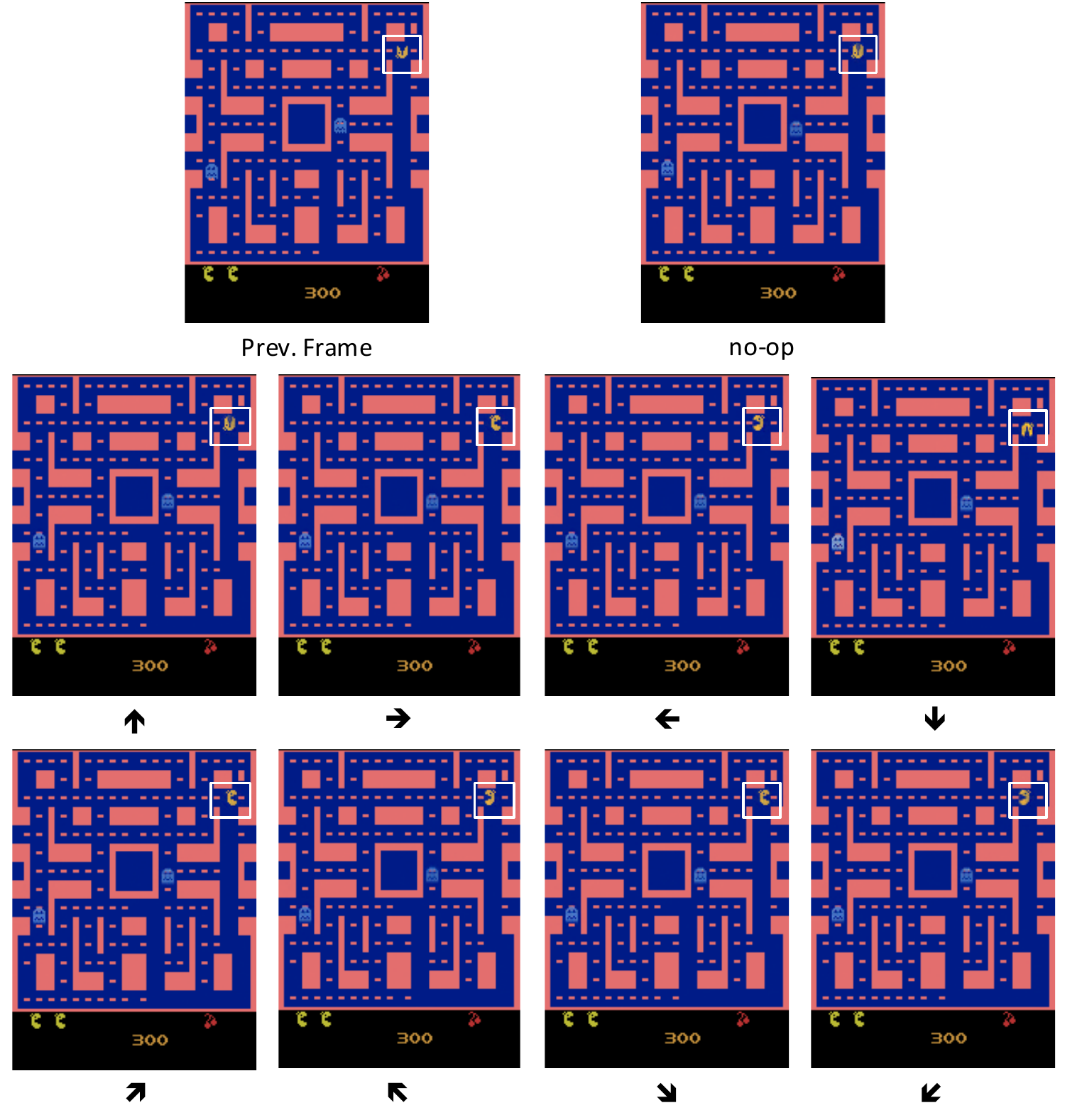} 
		\caption{Ms Pacman}
	\end{subfigure}
\caption{Predictions given different actions} 
\end{figure}

\section{Prediction Video}
\vspace{-10pt}
\begin{figure}[H]
	\centering
	\begin{subfigure}{\textwidth}
		\includegraphics[width=1.0\linewidth]{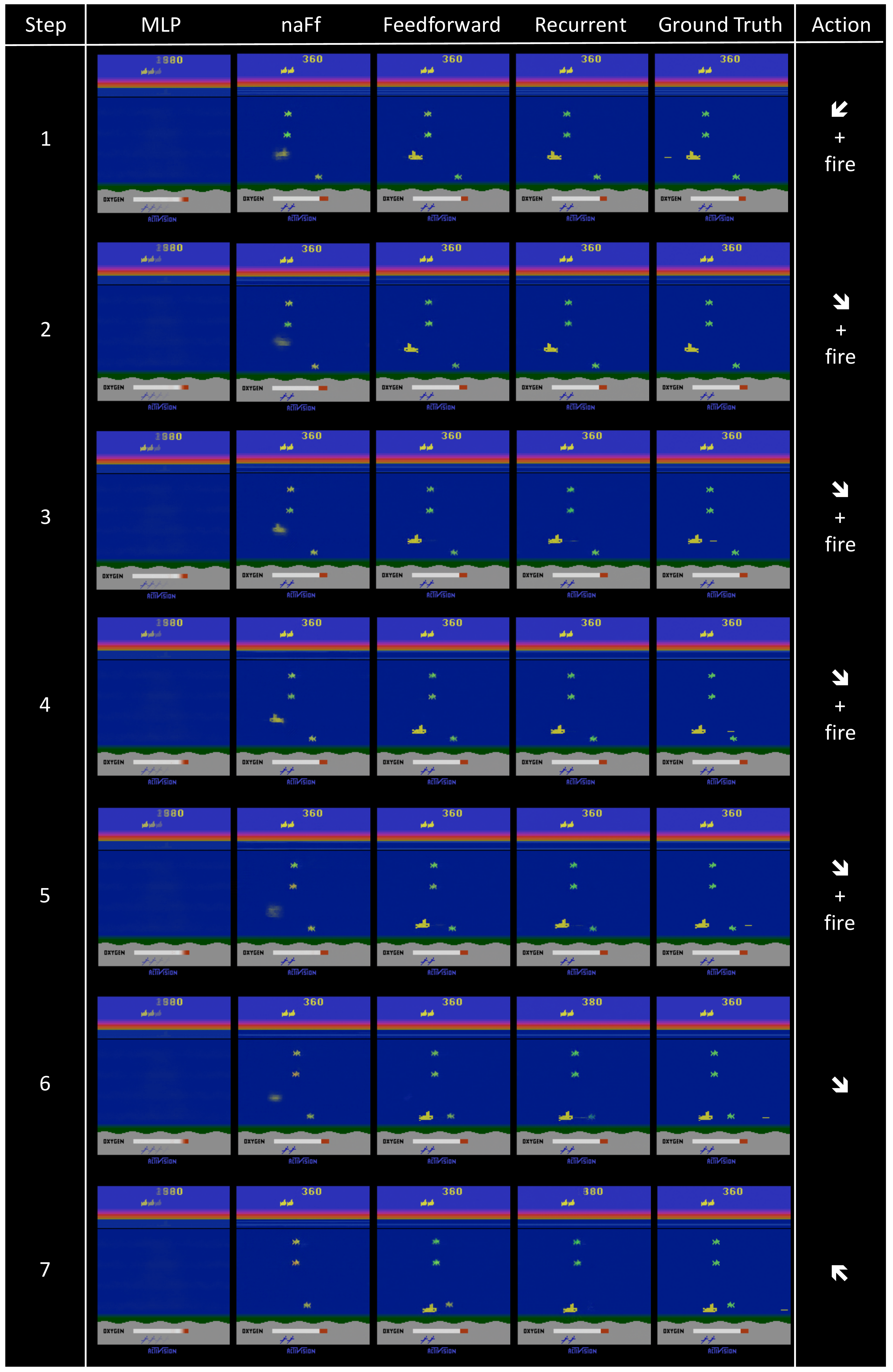} 
		\caption{Seaquest ($1\sim7$ steps). Our models predict the movement of the enemies and the yellow submarine which is controlled by actions. `naFf' predicts only the movement of other objects correctly, and the submarine disappears after a few steps. `MLP' does not predict any objects but generate only the mean pixel image. }
	\end{subfigure}
\end{figure}
\clearpage
\begin{figure}[H]
	\centering
	\ContinuedFloat
	\begin{subfigure}{\textwidth}
		\includegraphics[width=1.0\linewidth]{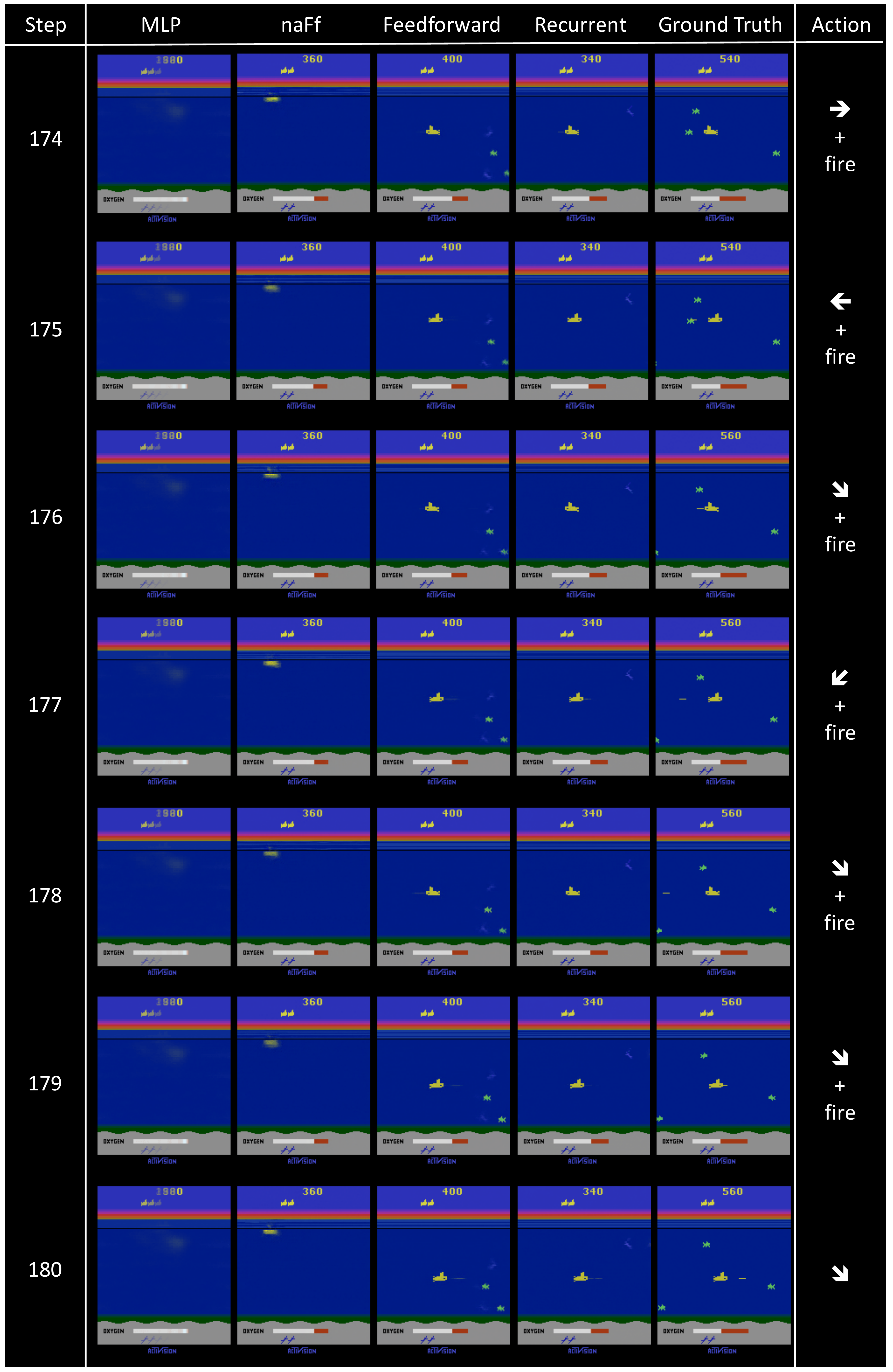} 
		\caption{Seaquest ($174\sim180$ steps). The proposed models predict the location of the controlled object accurately over 180-step predictions. They generate new objects such as fishes and human divers. Although the generated objects do not match the ground-truth images, their shapes and colors are realistic. }
	\end{subfigure}
\end{figure}
\clearpage
\begin{figure}[H]
	\centering
	\ContinuedFloat
	\begin{subfigure}{\textwidth}
		\includegraphics[width=1.0\linewidth]{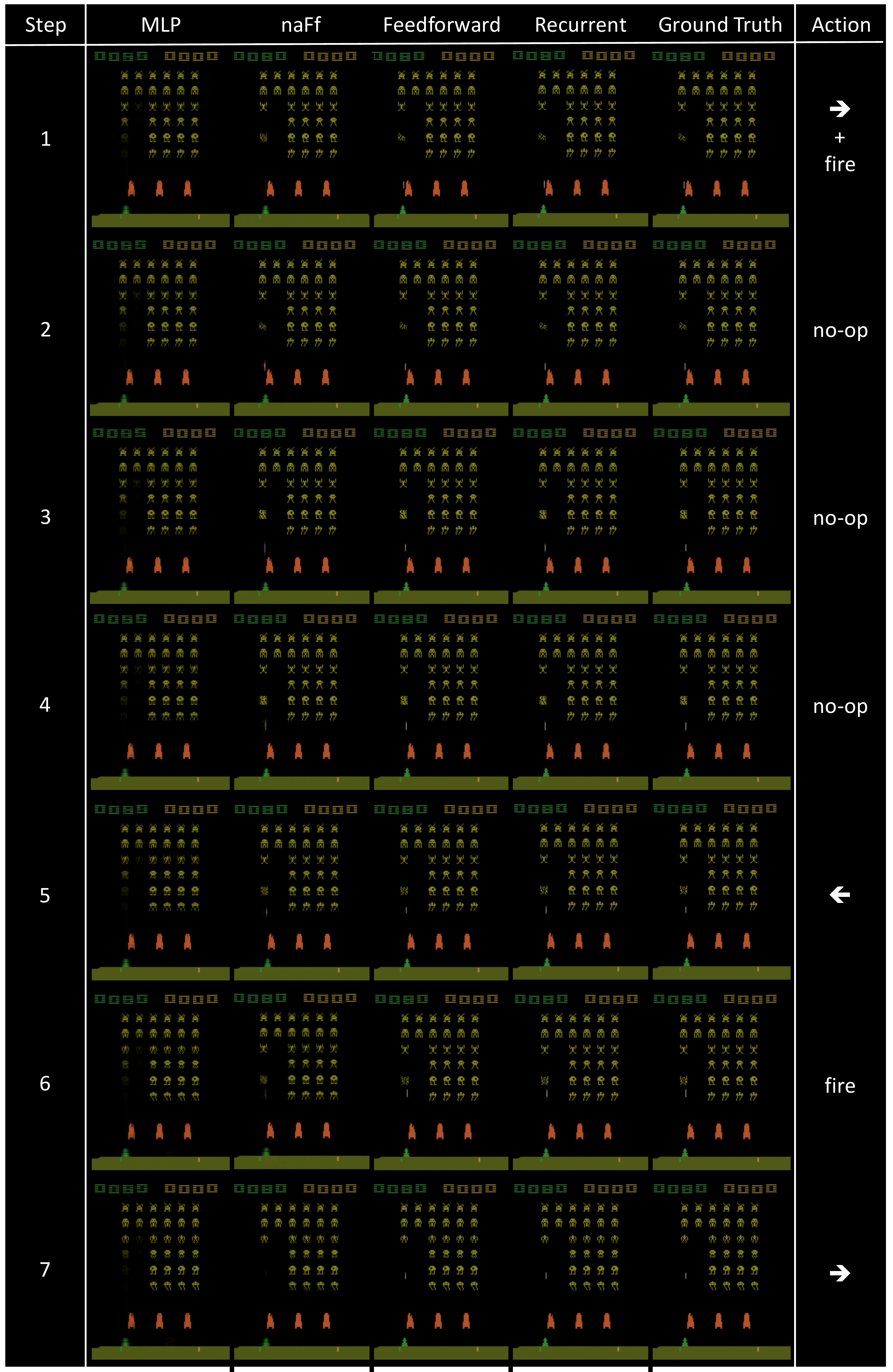} 
		\caption{Space Invaders ($1\sim7$ steps). The enemies in the center move and change their shapes from step $6$ to step $7$. This movement is predicted by the proposed models and `naFf', while the predictions from `MLP' are almost same as the last input frame. }
	\end{subfigure}
\end{figure}
\clearpage
\begin{figure}[H]
	\centering
	\ContinuedFloat
	\begin{subfigure}{\textwidth}
		\includegraphics[width=1.0\linewidth]{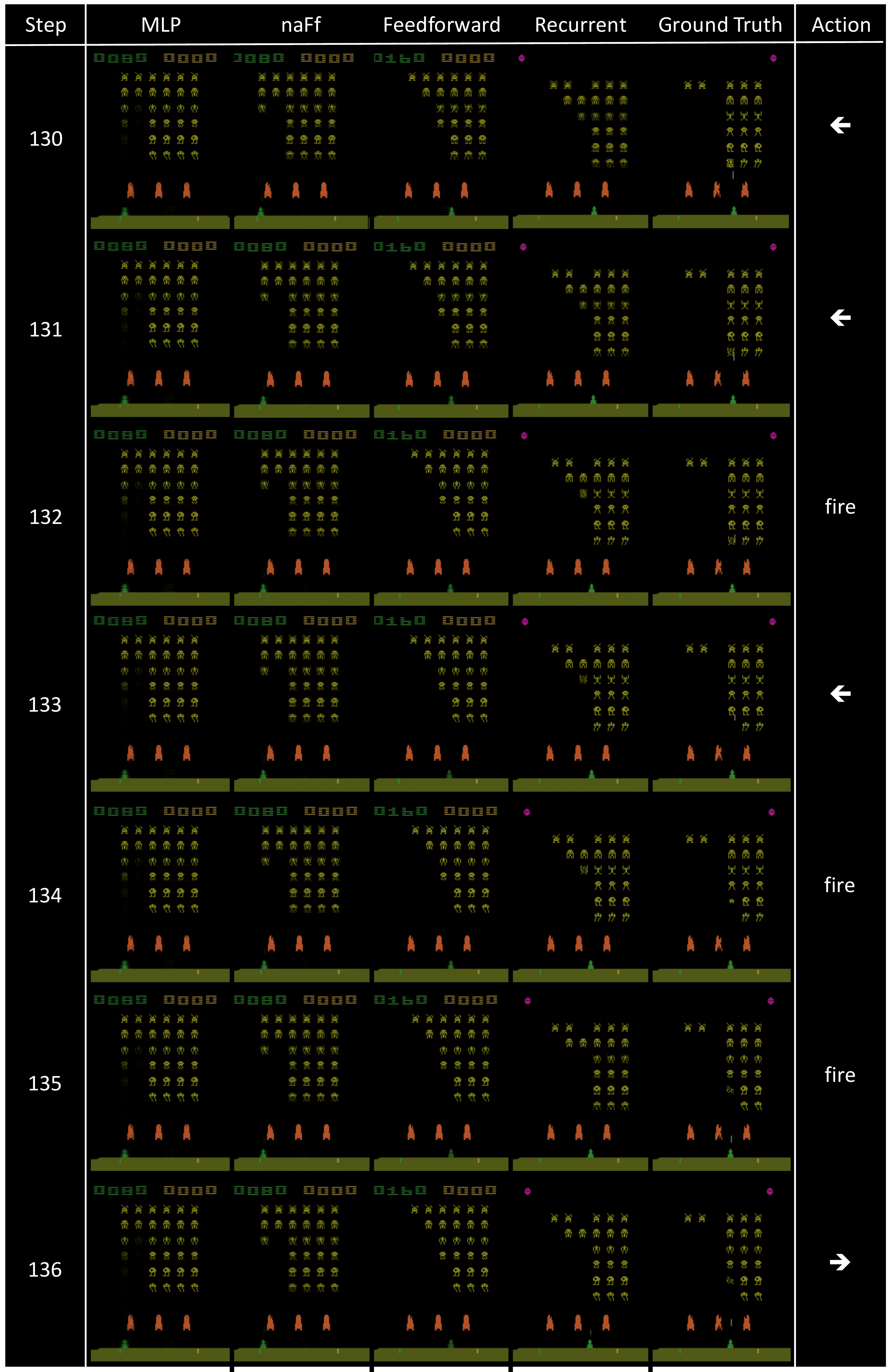} 
		\caption{Space Invaders ($130\sim136$ steps). Although our models make errors in the long run, the generated images are still realistic in that the objects are reasonably arranged and moving in the right directions. On the other hand, the frames predicted by `MLP' and `naFf' are almost same as the last input frame. }
	\end{subfigure}
\end{figure}
\clearpage
\begin{figure}[H]
	\centering
	\ContinuedFloat
	\begin{subfigure}{\textwidth}
		\includegraphics[width=1.0\linewidth]{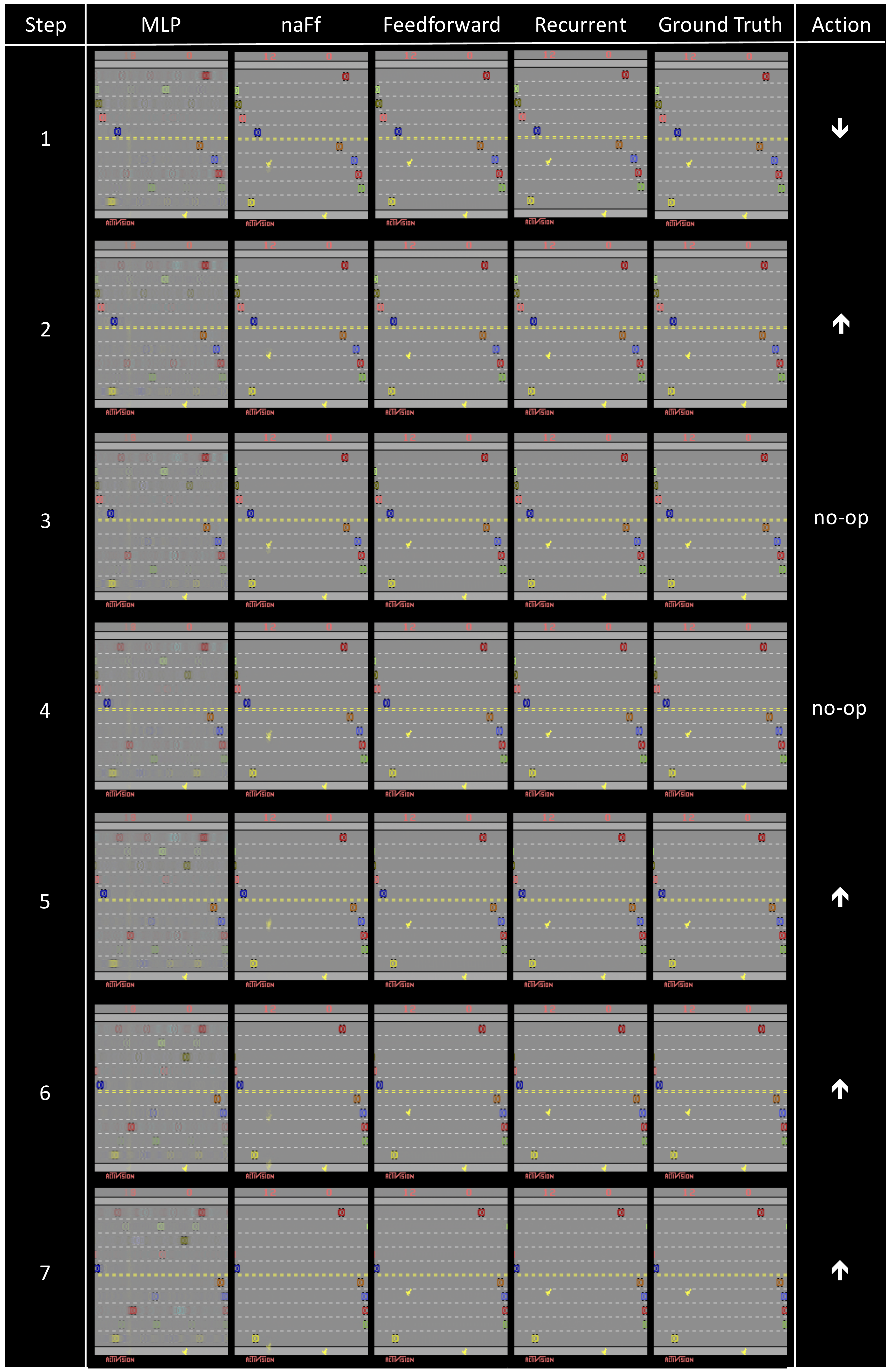} 
		\caption{Freeway ($1\sim7$ steps). The proposed models predict the movement of the controlled object correctly depending on different actions, while `naFf' fails to handle different actions. `MLP' generates blurry objects that are not realistic. }
	\end{subfigure}
\end{figure}
\clearpage
\begin{figure}[H]
	\centering
	\ContinuedFloat
	\begin{subfigure}{\textwidth}
		\includegraphics[width=1.0\linewidth]{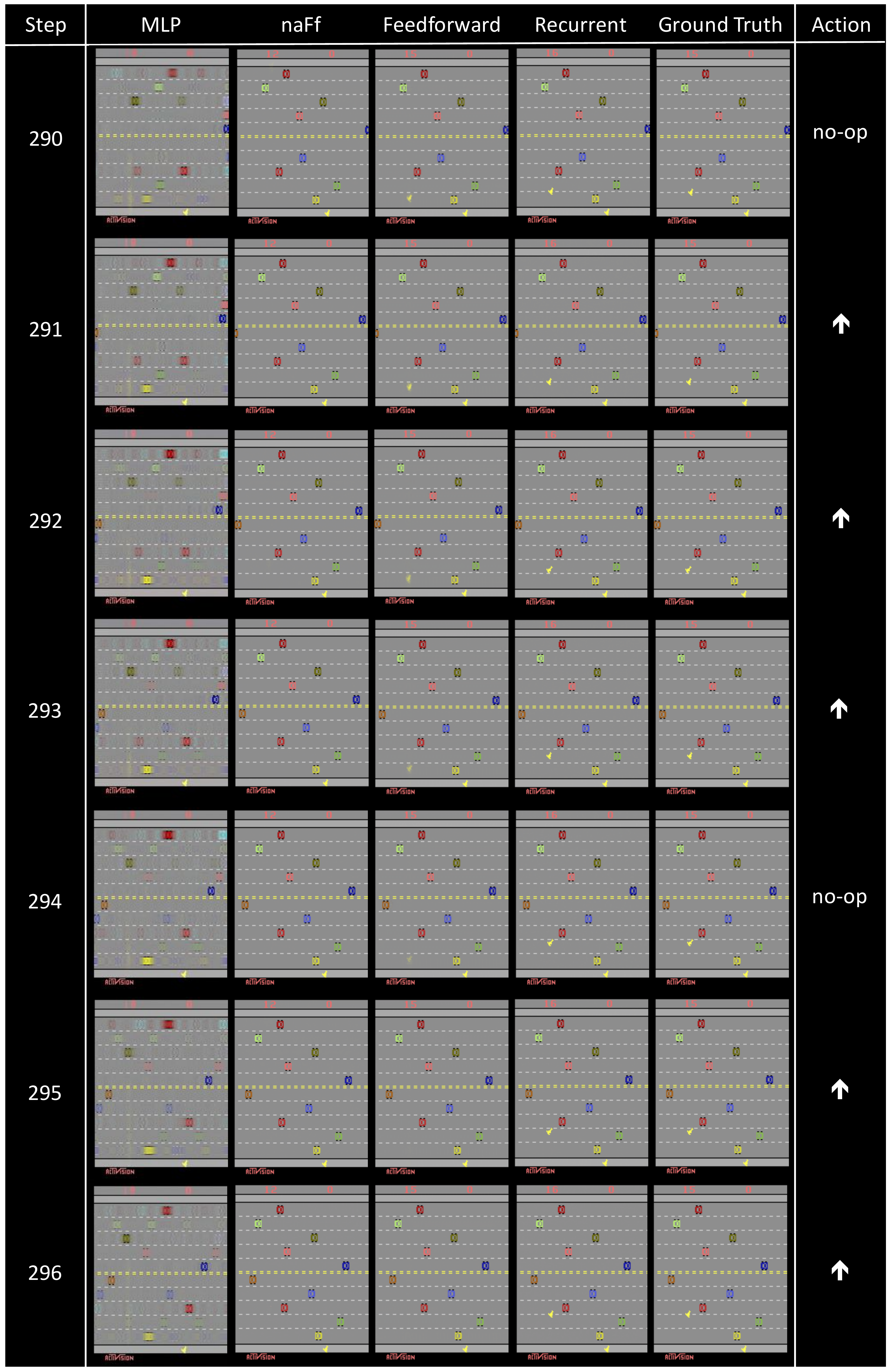} 
		\caption{Freeway ($290\sim296$ steps). The feedforward network diverges at 294-step as the agent starts a new stage from the bottom lane. This is due to the fact that actions are ignored for 9-steps when a new stage begins, which is not successfully handled by the feedforward network. }
	\end{subfigure}
\end{figure}
\clearpage
\begin{figure}[H]
	\centering
	\ContinuedFloat
	\begin{subfigure}{\textwidth}
		\includegraphics[width=1.0\linewidth]{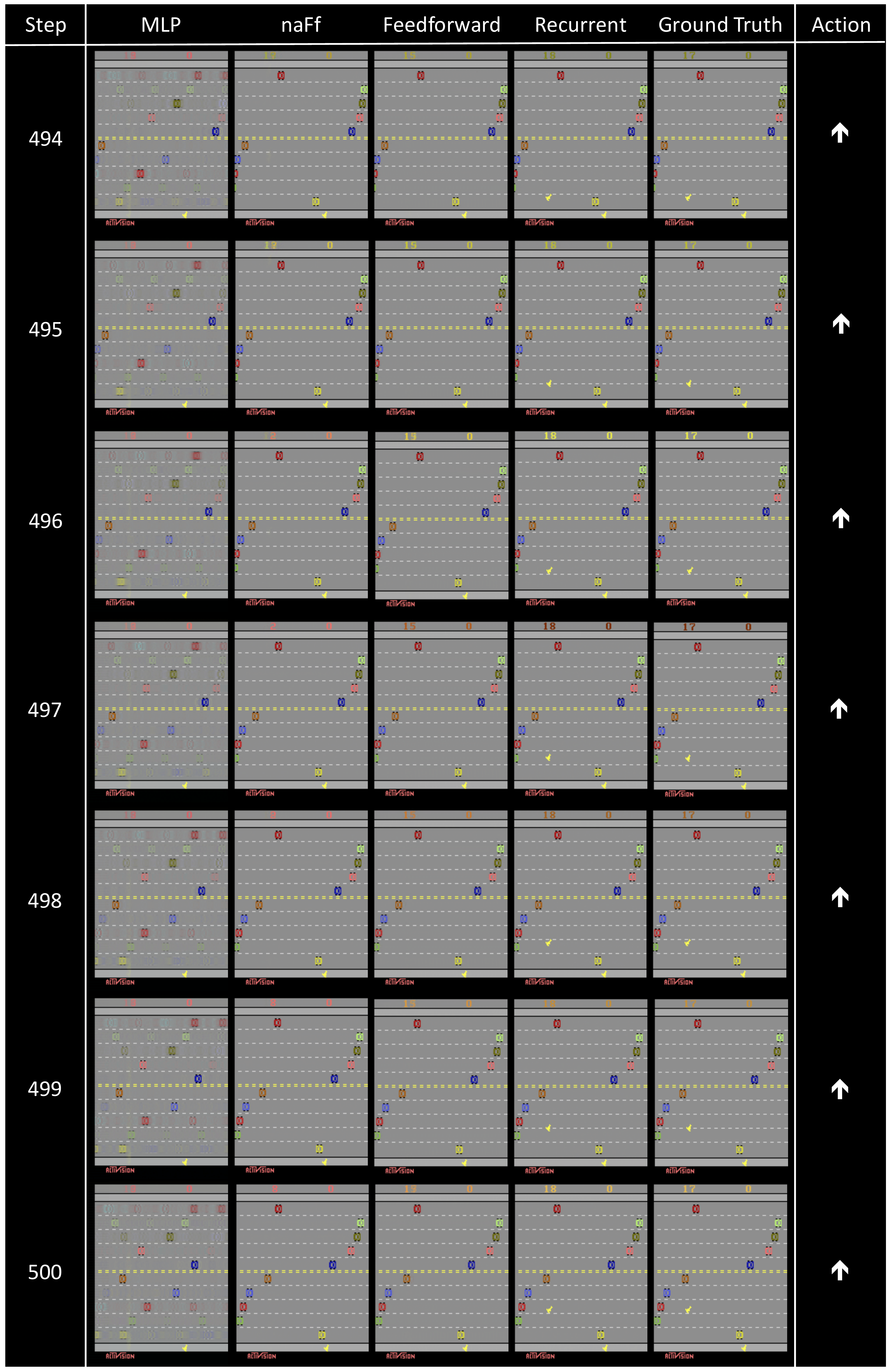} 
		\caption{Freeway ($494\sim500$ steps). The recurrent encoding model keeps track of every object over 500 steps. }
	\end{subfigure}
\end{figure}
\clearpage
\begin{figure}[H]
	\centering
	\ContinuedFloat
	\begin{subfigure}{\textwidth}
		\includegraphics[width=1.0\linewidth]{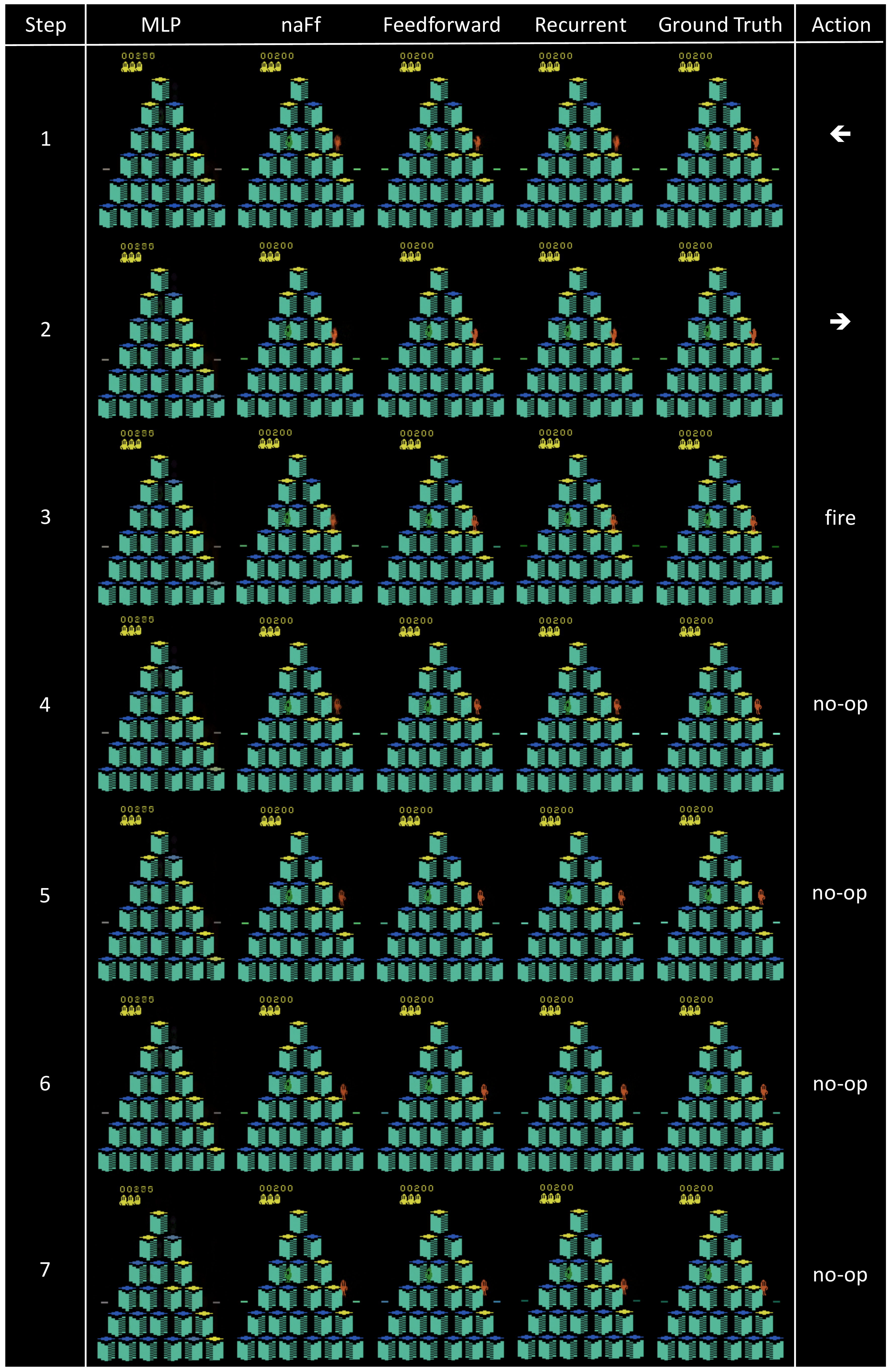} 
		\caption{QBert ($1\sim7$ steps). The controlled object jumps from the third row to the fourth row. In the meantime (jumping), the actions chosen by the agent do not have any effects. Our models and `naFf' predicts this movement, whereas `MLP' does not predict any objects. }
	\end{subfigure}
\end{figure}
\clearpage
\begin{figure}[H]
	\centering
	\ContinuedFloat
	\begin{subfigure}{\textwidth}
		\includegraphics[width=1.0\linewidth]{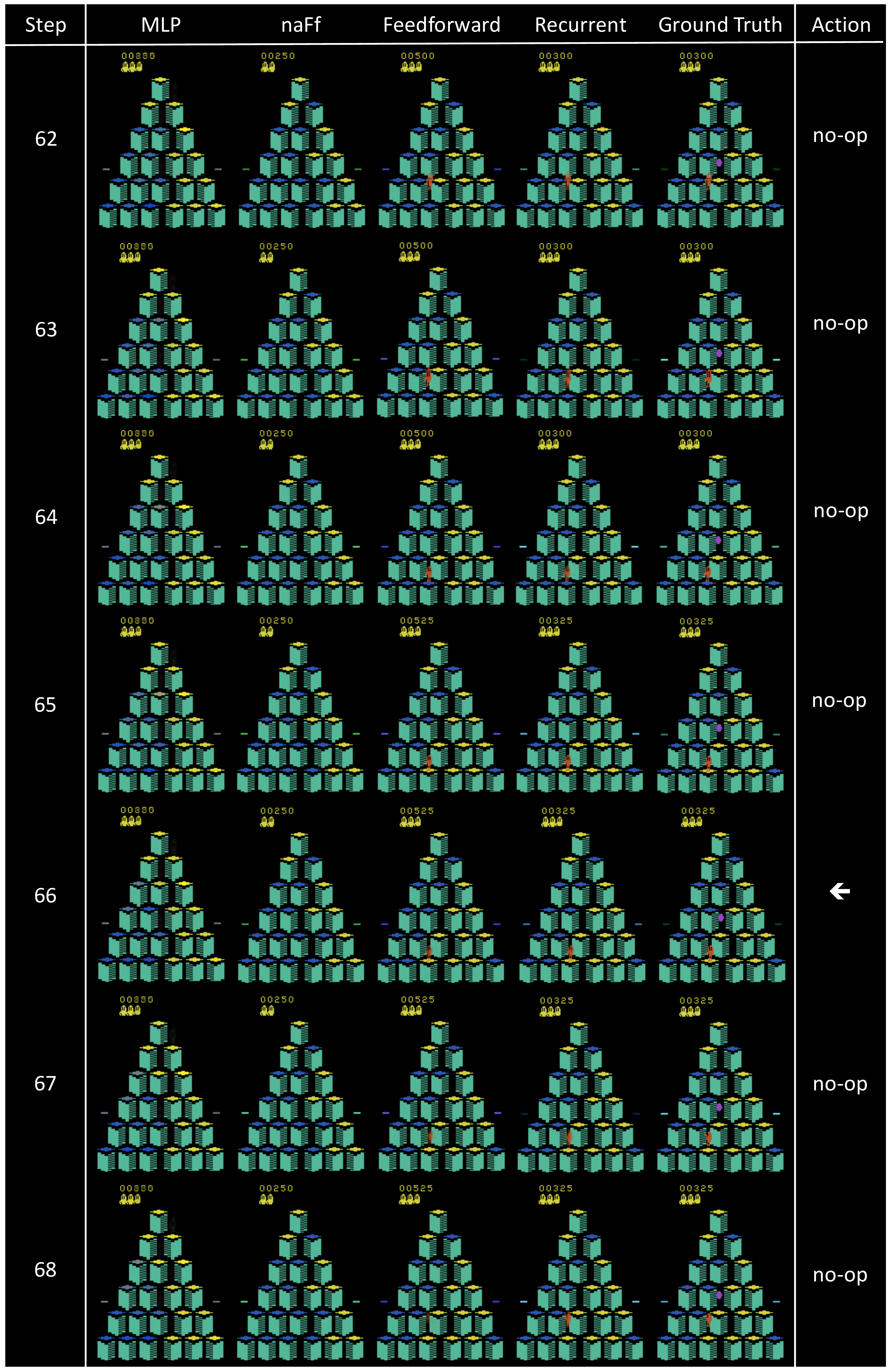} 
		\caption{QBert ($62\sim68$ steps). The recurrent model predicts the controlled object and the color of the cubes correctly, while the feedforward model diverges at 68-step as it predicts a blurry controlled object at 66-step. The baselines diverged before 62-step. }
	\end{subfigure}
\end{figure}
\clearpage
\begin{figure}[H]
	\centering
	\ContinuedFloat
	\begin{subfigure}{\textwidth}
		\includegraphics[width=1.0\linewidth]{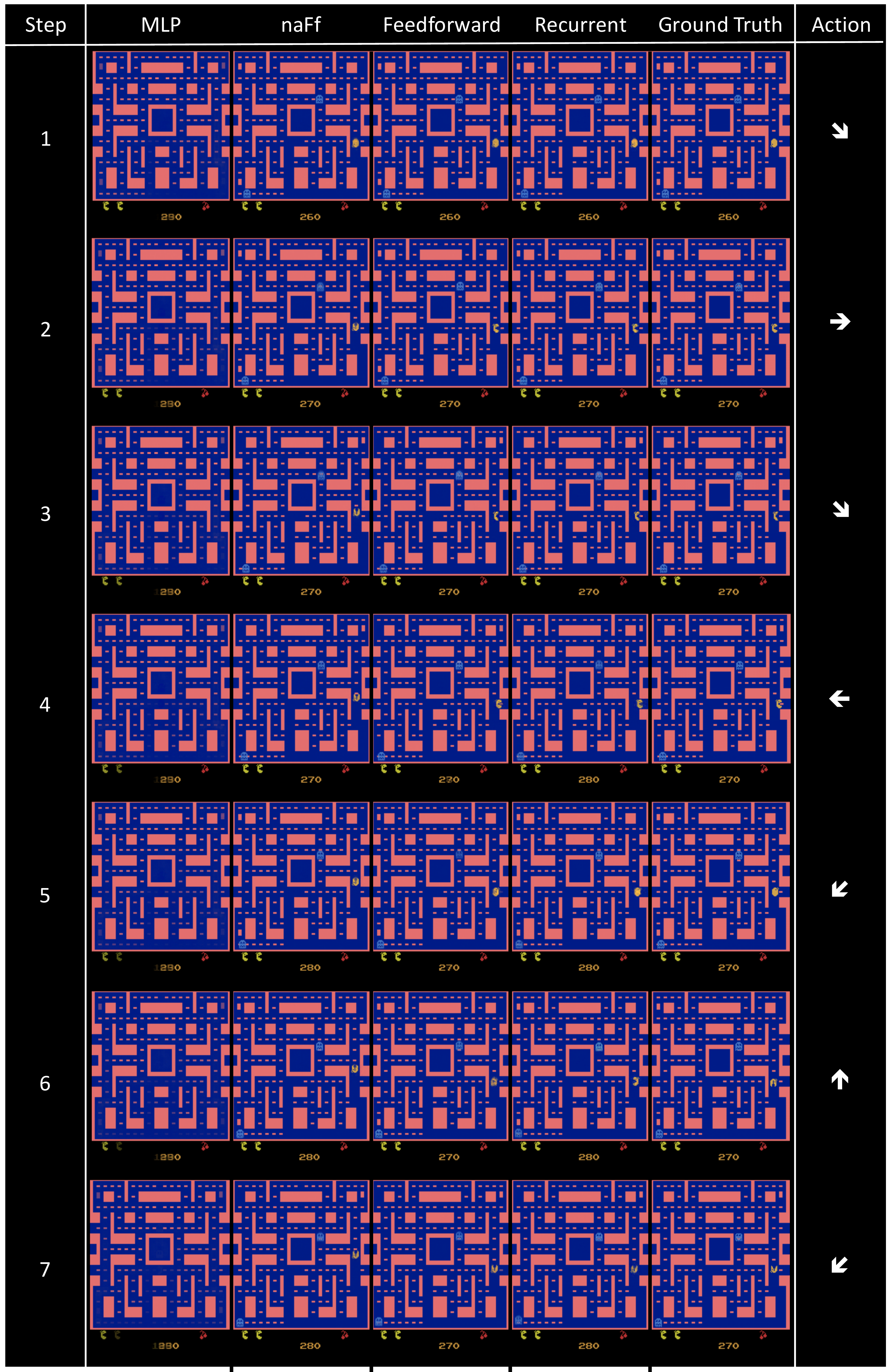} 
		\caption{Ms Pacman ($1\sim7$ steps). The proposed models predict different movements of Pacman depending on different actions, whereas `naFf' ignores the actions. `MLP' predicts the mean pixel image. }
	\end{subfigure}
\end{figure}
\clearpage
\begin{figure}[H]
	\centering
	\ContinuedFloat
	\begin{subfigure}{\textwidth}
		\includegraphics[width=1.0\linewidth]{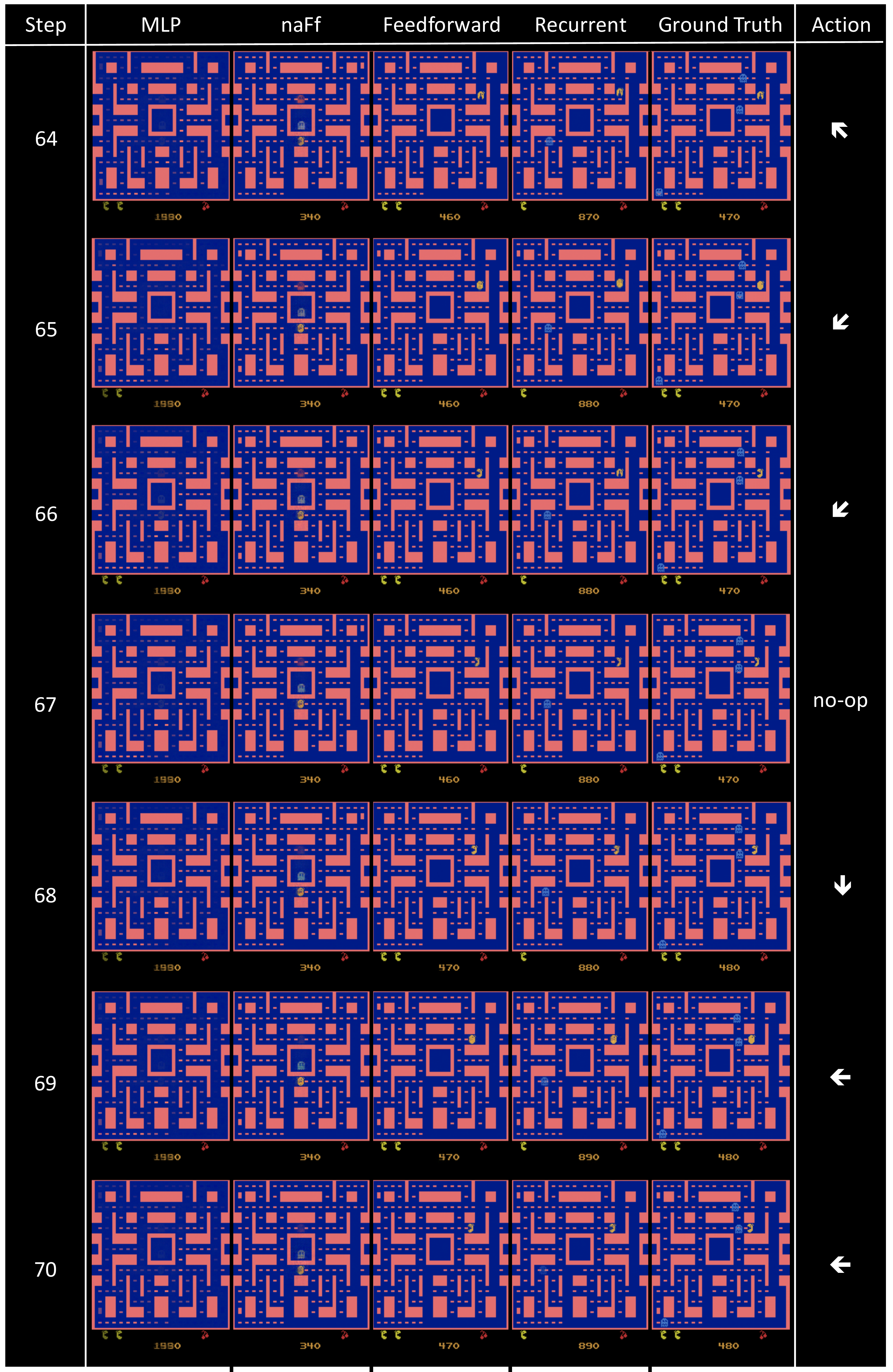} 
		\caption{Ms Pacman ($64\sim70$ steps). Our models keep track of Pacman, while they fail to predict the other objects that move almost randomly. }
	\end{subfigure}
\end{figure}
\end{document}